\documentclass{article}

\usepackage{microtype}
\usepackage{graphicx}
\usepackage{subfigure}
\usepackage{booktabs} 
\usepackage{amsmath}
\usepackage{amssymb}
\usepackage{amsfonts}
\usepackage{mathtools}
\usepackage{multirow}

\usepackage{hyperref}

\usepackage[accepted]{icml2020}

\DeclareMathOperator*{\argmax}{argmax}

\begin{document}

\twocolumn[
\icmltitle{Confidence-Aware Learning for Deep Neural Networks}

\icmlsetsymbol{equal}{*}

\begin{icmlauthorlist}
\icmlauthor{Jooyoung Moon}{equal,ds}
\icmlauthor{Jihyo Kim}{equal,ds}
\icmlauthor{Younghak Shin}{lgcns}
\icmlauthor{Sangheum Hwang}{ds,iise}
\end{icmlauthorlist}

\icmlaffiliation{ds}{Department of Data Science, Seoul National University of Science and Technology, Seoul, Republic of Korea}
\icmlaffiliation{lgcns}{LG CNS, Seoul, Republic of Korea}
\icmlaffiliation{iise}{Department of Industrial \& Information Systems Engineering, Seoul National University of Science and Technology, Seoul, Republic of Korea}

\icmlcorrespondingauthor{Sangheum Hwang}{shwang@seoultech.ac.kr}

\icmlkeywords{Confidence Estimation, Uncertainty Estimation, Deep Learning}

\vskip 0.3in
]

\printAffiliationsAndNotice{\icmlEqualContribution}

\begin{abstract}
    Despite the power of deep neural networks for a wide range of tasks, an overconfident prediction issue has limited their practical use in many safety-critical applications. Many recent works have been proposed to mitigate this issue, but most of them require either additional computational costs in training and/or inference phases or customized architectures to output confidence estimates separately. In this paper, we propose a method of training deep neural networks with a novel loss function, named \textit{Correctness Ranking Loss}, which regularizes class probabilities explicitly to be better confidence estimates in terms of ordinal ranking according to confidence. The proposed method is easy to implement and can be applied to the existing architectures without any modification. Also, it has almost the same computational costs for training as conventional deep classifiers and outputs reliable predictions by a single inference. Extensive experimental results on classification benchmark datasets indicate that the proposed method helps networks to produce well-ranked confidence estimates. We also demonstrate that it is effective for the tasks closely related to confidence estimation, out-of-distribution detection and active learning.
\end{abstract}

\section{Introduction}
\label{introduction}

Deep neural networks have shown remarkable performance on a wide spectrum of machine learning tasks for a variety of domains, e.g., image classification \citep{alex2012}, speech recognition \citep{hinton2012}, and medical diagnosis \citep{nam2019}. They are, however, generally an overconfident estimator that produces predictive probabilities with high confidence even for incorrect predictions \citep{nguyen2015,szegedy2014}.

The overconfident prediction issue makes deep neural network models unreliable, and therefore limits the deployment of the models in safety-critical applications such as autonomous driving and computer-aided diagnosis. For the successful integration of a deep neural network model into real-world systems, the model must not only be accurate but also indicate when it is likely to be wrong. In other words, \emph{the model should know what it does not know}. Hence, a deep neural network model that provides high quality of confidence estimates is required for practical applications. 

The quality of confidence estimates associated with a model's prediction can be assessed in two separate perspectives: confidence calibration and ordinal ranking according to confidence values. Confidence calibration is the problem of predicting probability estimates that reflects the true correctness likelihood. Thus, a well-calibrated classifier outputs predictive probabilities that can be directly interpreted as predictions' confidence level. It is known that modern neural networks generate miscalibrated outputs in spite of their high accuracy \citep{guo2017}. \citet{guo2017} examined which factors influence calibration performances of deep neural networks and showed that temperature scaling, a simple post-processing technique to learn a single corrective constant, is very effective at calibrating a model's predictions. Obviously, confidence calibration alone is insufficient to evaluate the quality of predictive confidence since it is orthogonal to both classification accuracy and ranking performance \citep{kumar2019}. For instance, we can have a perfectly calibrated classifier if it outputs the probability of 0.5 on the two-class dataset consisting of 50\% positive and 50\% negative samples. It means that a well-calibrated model may show lower predictive performances \citep{guo2017,neumann2018}.

Another view is ordinal ranking of predictions according to their confidence. Intuitively, a prediction with higher confidence value should be more likely to be correct than one with lower confidence value. Thus, ordinal ranking aims to estimate confidence values whose ranking among samples are effective to distinguish correct from incorrect predictions. In most previous studies, this problem has been casted into different tasks such as failure prediction \citep{hecker2018,jiang2018,corbiere2019}, selective classification \citep{el-yaniv2010,geifman2017}, and out-of-distribution detection \citep{devries2018,liang2018,lee2018_mahal,hendrycks2017,roady2019}, although they have tried to solve fundamentally similar problem under slightly different settings. A model that outputs well-ranked confidence estimates should work well on all of these tasks. In many practical settings, ordinal ranking performance is important since it is very closely related to measure whether the model knows what it knows. In this work, we focus on how to obtain good predictions in terms of ordinal ranking of confidence estimates.

Our goal is to learn a deep neural network for classification that outputs better predictive probabilities to quantify confidence values. In a classification problem, predictive probabilities by themselves must represent confidence estimates of predictions since a conditional distribution of classes given an input is assumed to be a multinomial distribution. With these probabilities, confidence estimates associated with them can be naturally computed by basic metrics including the maximum class probability (i.e., the largest softmax score), entropy, margin, etc., as commonly used to estimate uncertainty \citep{settles2009}. In other words, predictive probabilities from a \textit{well-trained} classifier are the essential ingredients to obtain confidence estimates of high quality. 

To build such a well-trained model, we propose a simple but effective regularization method that enforces a model to learn an ordinal ranking relationship. The proposed regularization method can be simply implemented via a ranking loss named \emph{Correctness Ranking Loss} (CRL) that incorporates a comparison of randomly selected a pair of samples. It is minimized when confidence estimates of samples with high probabilities of being correct are greater than those of samples with low probabilities of being correct. The main advantage of the proposed method is its computational efficiency, i.e., we need to compute just an additional loss value during training and can obtain high quality of confidence estimates by a single inference. Therefore, it can be universally applied to any architecture with little increase in computational costs.\footnote{In practice, the amount of computation for calculating loss can be completely negligible.}

We validate the proposed method through extensive experiments over various benchmark datasets for image classification and several popular architectures. The experimental results demonstrate that training with CRL is very effective to obtain well-ranked confidence estimates compared with existing methods specially designed to estimate them. With these well-ranked confidence estimates, it is also shown that a classifier alone works surprisingly well on other complicated tasks such as out-of-distribution (OOD) detection and active learning in which ordinal ranking of confidence is important.

\begin{figure*}[ht!]
    \vskip 0.0in
    \includegraphics[width=\textwidth]{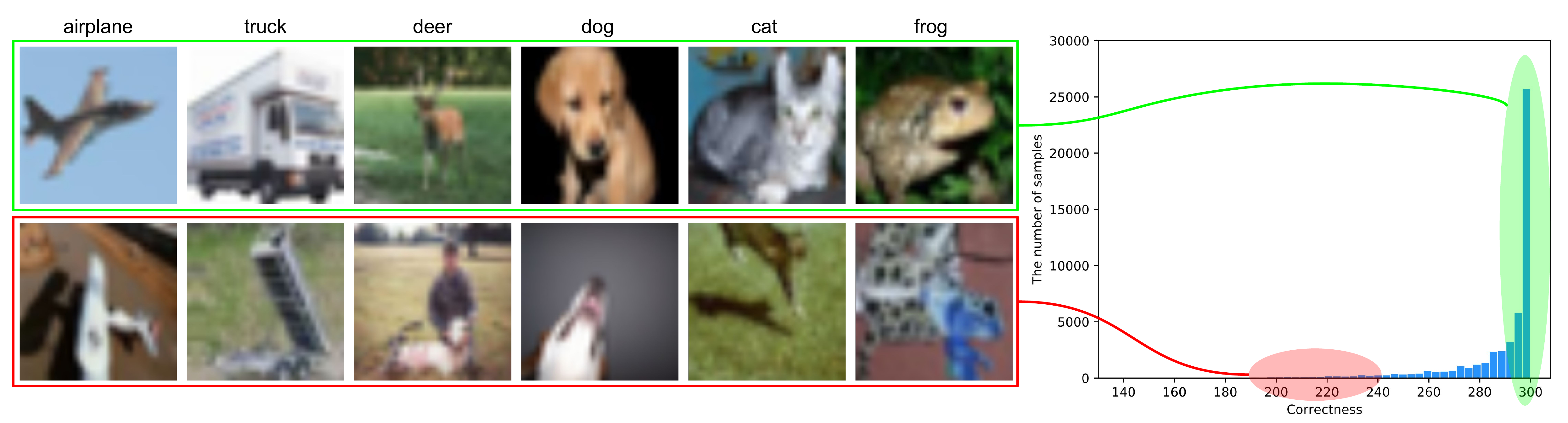}
    \vspace*{-7mm}
    \caption{Examples from CIFAR-10 whose number of correct prediction events are high (\textit{green group, top}) and low (\textit{red group, bottom}). The green group consists of easy-to-classify examples while the examples in the red group are hard to recognize.}
    \label{fig-motivation}
\end{figure*}

\section{Related Work} 
\label{related-work}

Confidence (or its opposite, uncertainty) estimation in predictions with modern neural networks becomes actively studied in the machine learning community. Bayesian approach provides a natural representation of uncertainty in a neural network by allowing rich probabilistic interpretations for a model's predictions. With a prior distribution over model parameters of a neural network, several approximate Bayesian methods can be employed to infer the posterior distribution over the parameters which accounts for predictive uncertainty, for instance, Laplace approximation \citep{mackay1992}, Markov Chain Monte Carlo (MCMC) \citep{neal1996} and variational inference \citep{graves2011}. While these methods are effective for small neural networks, it is computationally expensive for modern deep neural networks.
In the study of \citet{gal2016}, they proposed Monte Carlo dropout (MCdropout) that uses dropout \citep{srivastava2014} at test time to estimate predictive uncertainty by sampling several stochastic predictions. It has gained attention as a practical approximate Bayesian method for uncertainty estimation \citep{gurau2018,zhang2019}. 
\citet{kendall2017} presented a framework to decompose the uncertainty into aleatoric one capturing noise inherent in the data and epistemic one accounting for the model's uncertainty.
Although they greatly reduce computational costs for estimating uncertainty, it still requires multiple forward passes for inference.

As another line of study, there are also several works based on non-Bayesian approach to obtain confidence estimates. In standard deep classifiers, predictive class probabilities that can be used for confidence estimation are naturally appeared as softmax outputs. \citet{hendrycks2017} showed a simple threshold-based method utilizing confidence estimates from softmax scores are quite effective for both ordinal ranking and OOD detection tasks. \citet{liang2018} introduced an OOD detector named ODIN to improve the OOD detection performances by applying temperature scaling and adding small perturbations to inputs, and 
\citet{lee2018_mahal} proposed a confidence estimation method using the Mahalanobis distance on feature spaces of deep classifiers which can be further enhanced in conjunction with input perturbations and feature ensembling.  
Except for \citet{hendrycks2017}, they are designed specifically for the OOD detection task, and none of the ordinal ranking performances was reported. 

Some recent studies have tried to directly learn confidence estimates with deep classifiers by augmenting a network's architecture \citep{devries2018,corbiere2019}. Specifically, they have an additional output node that produces confidence estimates and utilize these estimates for OOD detection \citep{devries2018} or ordinal ranking \citep{corbiere2019}. However, they rely on the predictive performance of confidence estimates from the node since confidence estimates are generated independently of class probabilities. 

Ensembles of neural networks, simply called Deep Ensembles, are certainly useful for confidence estimation as can be seen in \citet{lakshminarayanan2017}. \citet{geifman2018} found that the confidence estimates of easily learnable samples become impaired during training process with a stochastic gradient descent (SGD) based optimizer. To address this issue, they suggest using the Average Early Stopping (AES) algorithm similar to the snapshot ensembles \citep{huang2017} to leverage the quality of confidence estimates in terms of ordinal ranking. However, these approaches are inherently computationally demanding.

Compared to the previous studies, our proposed method neither increases computational costs for training and inference nor augments architectures of standard deep classifiers to have good confidence estimates. With the proposed method, a standard classification network such as examined in \citet{hendrycks2017} can become a very strong baseline that yields much better confidence estimates.

\section{Confidence-Aware Learning}

In this section, we introduce the ordinal ranking problem and empirical findings that motivates our work. Then, we provide in-depth descriptions of the proposed \emph{Correctness Ranking Loss} with implementation details.

\subsection{Problem Statement}
\label{sec-prob-stmt}

In this work, we address a multi-class classification problem with a deep neural network that utilizes a standard softmax layer to output predicted class probabilities. 

Let $\mathcal{D}=\{(\mathbf{x}_i, y_i)\}_{i=1}^n$ be a dataset consisting of $n$ labeled samples from a joint distribution over $\mathcal{X} \times \mathcal{Y}$ where $\mathcal{X}$ is an input space and $\mathcal{Y}=\{1,2,3,..,K\}$ is a label set for the classes.
A deep neural classifier $f$ is a function $f:\mathcal{X} \rightarrow \mathcal{Y}$ that produces the predicted class probabilities $\mathbf{p}_i=P\left( y\middle|\mathbf{x}_i, \mathbf{w} \right)$ for a sample $i$ where $\mathbf{w}$ is a set of model parameters of the network. With these probabilities, the predicted class $\hat{y}_i$ of an input $\mathbf{x}_i$ is determined as $\hat{y}_i = \argmax_{y \in \mathcal{Y}} P( y|\mathbf{x}_i, \mathbf{w} )$.

From the predicted class probabilities computed by a softmax layer, we can have several confidence estimates: a class probability associated with $\hat{y}_i$ (i.e., the maximum class probability), negative entropy\footnote{For entropy, confidence should be the negative of entropy.}, and margin. Margin is defined as the difference between the predicted probabilities of the first and second most probable classes \citep{settles2009}.

Ordinal ranking, also known as failure prediction \citep{corbiere2019} or error detection \citep{hendrycks2017}, is the problem about ranking among samples to distinguish correct from incorrect predictions according to their confidence estimates. In case of perfect ordinal ranking, every pair of $(\mathbf{x}_i,y_i)$ and $(\mathbf{x}_j,y_j)$ from the true joint distribution should hold the following relationship:
\begin{equation}
\label{eq-ordinal-ranking}
\begin{split}
    &\kappa(\mathbf{p}_i|\mathbf{x}_i, \mathbf{w}) \leq \kappa(\mathbf{p}_j|\mathbf{x}_j, \mathbf{w}) \\
    \iff &P( \hat{y}_i=y_i|\mathbf{x}_i ) \leq P( \hat{y}_j=y_j|\mathbf{x}_j )
\end{split}
\end{equation}
where $\kappa$ denotes a confidence function (e.g., the maximum class probability, negative entropy, and margin). Note that $P( \hat{y}_i=y_i|\mathbf{x}_i)$ represents the true probability of being correct for a sample $i$. It is desirable for a model to learn the relationship in Eq.~\eqref{eq-ordinal-ranking} during training.

\subsection{Motivation}
\label{sec-motivation}

Ideally we expect that a model can learn the relationship in Eq.~\eqref{eq-ordinal-ranking} directly during training. However, estimating the true probability of getting a correct prediction is the major obstacle. It is generally impractical since we do not know the true joint distribution over $\mathcal{X} \times \mathcal{Y}$ and a classifier $f$ is gradually biased towards the training dataset as training proceeds.

We hypothesis that the probability of being correct is \textit{roughly} proportional to the frequency of correct predictions during training with SGD-based optimizers. The empirical findings in \citet{toneva2019} and \citet{geifman2018} support our hypothesis. \citet{toneva2019} investigated the number of forgetting events for each sample and showed that samples being frequently forgotten are relatively more difficult to classify. Similarly, \citet{geifman2018} observed that easy-to-classify samples are learned earlier during training compared to hard-to-classify samples. Motivated by these findings, we expect that the frequency of correct prediction events for each sample examined on SGD-based optimizer's trajectory can be used as a good proxy for the probability estimates of being correct for it.\footnote{Strictly speaking, this is not an appropriate estimator of probability in a statistical sense since correct prediction events are not \textit{i.i.d.} observations.}  

Figure~\ref{fig-motivation} shows the distribution of correct prediction events for training data and examples sampled according to their number of correct prediction events. For this visual inspection, we trained PreAct-ResNet110 \citep{he2016} on CIFAR-10 dataset \citep{krizhevsky2009} for 300 epochs. To count the correct prediction events of each sample, we consider only samples in the current mini-batch, and therefore all samples are examined once per epoch. The top green box contains the images that are correctly classified with high frequency and the bottom red box consists of less correctly classified images. Examples in the green box contain the complete shape of objects with a clearly distinguishable background, and therefore they are easy to recognize. On the other hand, examples in the red box are hard to classify into true classes since the objects appear as a part or with other unrelated objects. Based on this observation, we suppose that it is able to estimate the probability of being classified correctly by the frequency of correct prediction events.

\subsection{Correctness Ranking Loss (CRL)}
\label{sec-crl}

It is enabled to design a loss function to reflect the desirable ordinal ranking of confidence estimates in Eq.~\eqref{eq-ordinal-ranking} if the true class probability is estimated by how many times a sample is classified correctly during training. The loss function should be affected by whether the ranking of a pair of samples is right or not, and the loss will be incurred when the relationship in Eq.~\eqref{eq-ordinal-ranking} is violated.

We propose CRL so that a classifier learns the ordinal ranking relationship. For a pair of $\mathbf{x}_i$ and $\mathbf{x}_j$, it is defined as
\begin{equation}
\label{eq-cr-loss}
    \mathcal{L}_{\text{CR}}(\mathbf{x}_i,\mathbf{x}_j) = \max(0, -g(c_i, c_j)(\kappa_i - \kappa_j) + |c_i - c_j|)
\end{equation}
where $c_i$ is the proportion of correct prediction events of $\mathbf{x}_i$ over the total number of examinations (i.e., $c_i \in [0,1]$), $\kappa_i$ represents $\kappa(\mathbf{p}_i|\mathbf{x}_i, \mathbf{w})$ and
\begin{equation*}
    g(c_i, c_j) = \begin{cases}
    1, & \text{if $c_i > c_j$} \\
    0, & \text{if $c_i = c_j$} \\
    -1, & \text{otherwise} \\
    \end{cases}
\end{equation*}
As can be seen in Figure~\ref{fig-motivation}, in general, the distribution of correct prediction events is highly skewed to the left especially for modern neural networks showing high performance. It means that most training samples are correctly classified during the whole course of training. For those samples, learning the ranking relationship is meaningless. Moreover, our model should learn more from a pair of samples with a large difference in $c$ values rather than those with a small difference. To this end, we introduce some margin $|c_i - c_j|$ to CRL. 
As a result, CRL enforces a model to output well-ranked $\kappa_i$'s. For example, for a pair with $c_i > c_j$, CRL will be zero when $\kappa_i$ is larger than $\kappa_j+|c_i - c_j|$, but otherwise a loss will be incurred. 
Given a labeled dataset $\mathcal{D}$, the total loss function $\mathcal{L}$ is a weighted sum of a cross-entropy loss $\mathcal{L}_{\text{CE}}$ and a CRL $\mathcal{L}_{\text{CR}}$:
\begin{equation}
\label{eq-total-loss}
    \mathcal{L} = ~~\sum_{\mathclap{(\mathbf{x}_i, y_i) \in \mathcal{D}}}~~{\mathcal{L}_{\text{CE}}(\mathbf{p}_i,y_i)} + \lambda 
    ~~\sum_{\mathclap{(\mathbf{x}_i,\mathbf{x}_j) \in \mathcal{D}_C}}~~{ 
    \mathcal{L}_{\text{CR}}(\mathbf{x}_i,\mathbf{x}_j) }
\end{equation}
where $\lambda$ is a constant controlling the influence of $\mathcal{L}_{\text{CR}}$ and $\mathcal{D}_C$ denotes a set of all possible sample pairs from $\mathcal{D}$.

\textbf{Implementation details.} 
With a mini-batch of size $b$, $\{(\mathbf{x}_{[i]}, y_{[i]})\}_{i=1}^b$, $\mathcal{L}_{\text{CR}}$ should be computed over all possible sample pairs at each model update. However, it is computationally expensive, so we employ a few approximation schemes following to \citet{toneva2019} for reducing the costs. First, only samples in the current mini-batch are examined to determine whether each sample is correctly classified or not as done in Section~\ref{sec-motivation}. Note that it can be judged by softmax outputs with no costs. Second, since the number of all possible pairs within a mini-batch is too large, we consider only $b$ pairs to include as many pairs of samples as the computational cost is manageable. Specifically, for $i=1,\dots,b-1$, $\mathbf{x}_{[i]}$ is paired with $\mathbf{x}_{[i+1]}$ and the last sample $\mathbf{x}_{[b]}$ is paired with $\mathbf{x}_{[1]}$. 

For the confidence function $\kappa$, we consider simple and popular three estimators: the maximum class probability, negative entropy, and margin. Confidence estimates from the maximum class probability and margin as well as $c_i$ naturally lies in $[0,1]$ while those from negative entropy does not. Therefore, confidence estimates obtained from negative entropy are normalized by using the min-max scaling.

\begingroup
\begin{table*}[ht]
    \centering
    \caption{Comparison of the quality of confidence estimates on various datasets and architectures. The means and standard deviations over five runs are reported. $\downarrow$ and $\uparrow$ indicate that lower and higher values are better respectively. For each experiment, the best result is shown in boldface. AURC and E-AURC values are multiplied by $10^3$, and NLL are multiplied by $10$ for clarity. All remaining values are percentage.}
    \label{table-clf-perf}
    \vskip 0.05in
    \resizebox{0.85\textwidth}{!}{
    \begin{tabular}{ll|ccccc|ccc}
\hline
\multirow{2}{*}{\textbf{\begin{tabular}[c]{@{}l@{}}\textbf{Dataset}\\\textbf{Model}\end{tabular}}} & \multirow{2}{*}{\textbf{\textbf{Method}}} & \multirow{2}{*}{\textbf{\begin{tabular}[c]{@{}c@{}}\textbf{\small{ACC}}\\ \small{($\uparrow$)}\end{tabular}}} & \multirow{2}{*}{\textbf{\begin{tabular}[c]{@{}c@{}}\textbf{\small{AURC}}\\ \small{($\downarrow$)}\end{tabular}}} & \multirow{2}{*}{\textbf{\begin{tabular}[c]{@{}c@{}}\textbf{\small{E-AURC}}\\ \small{($\downarrow$)}\end{tabular}}} & \multirow{2}{*}{\textbf{\begin{tabular}[c]{@{}c@{}}\textbf{\small{AUPR-}}\\ \textbf{\small{Err}} \footnotesize{($\uparrow$)}\end{tabular}}} & \multirow{2}{*}{\textbf{\begin{tabular}[c]{@{}c@{}}\textbf{\small{FPR-95\%}}\\ \textbf{\small{TPR}} \small{($\downarrow$)}\end{tabular}}} & \multirow{2}{*}{\begin{tabular}[c]{@{}c@{}}\textbf{\small{ECE}}\\ \small{($\downarrow$)}\end{tabular}} & \multirow{2}{*}{\begin{tabular}[c]{@{}c@{}}\textbf{\small{NLL}}\\ \small{($\downarrow$)}\end{tabular}} & \multirow{2}{*}{\begin{tabular}[c]{@{}c@{}}\textbf{\small{Brier}}\\ \small{($\downarrow$)}\end{tabular}} \\
 &  &  &  &  &  &  &  &  &  \\ \hline
\multirow{5}{*}{\begin{tabular}[c]{@{}l@{}}\textbf{CIFAR-10}\\ \textbf{VGG-16}\end{tabular}} & Baseline & 93.74$\pm$0.14 & 7.10$\pm$0.31 & 5.10$\pm$0.26 & 44.19$\pm$0.34 & 41.43$\pm$0.38 & 5.20$\pm$0.11 & 3.79$\pm$0.11 & 11.30$\pm$0.21 \\
 & AES($k$=10) & \textbf{93.97$\pm$0.12} & 7.15$\pm$0.25 & 5.30$\pm$0.25 & 44.47$\pm$1.00 & 41.01$\pm$1.75 & 1.61$\pm$0.27 & 2.06$\pm$0.04 & 9.26$\pm$0.15 \\
 & MCdropout & 93.78$\pm$0.27 & 6.72$\pm$0.28 & 4.72$\pm$0.19 & 45.08$\pm$2.14 & 41.52$\pm$2.83 & 1.11$\pm$0.19 & 1.93$\pm$0.05 & 9.34$\pm$0.39 \\
 & Aleatoric+MC & 93.91$\pm$0.13 & \textbf{6.57$\pm$0.29} & \textbf{4.68$\pm$0.22} & 44.67$\pm$1.76 & 41.68$\pm$1.86 & \textbf{0.86$\pm$0.12} & \textbf{1.89$\pm$0.05} & \textbf{9.08$\pm$0.24} \\
 & CRL-softmax & 93.82$\pm$0.18 & 6.78$\pm$0.18 & 4.83$\pm$0.08 & \textbf{46.79$\pm$1.75} & \textbf{40.21$\pm$2.18} & 1.24$\pm$0.20 & 2.09$\pm$0.04 & 9.33$\pm$0.21 \\ \hline
\multirow{5}{*}{\begin{tabular}[c]{@{}l@{}}\textbf{CIFAR-10}\\ \textbf{ResNet110}\end{tabular}} & Baseline & 94.11$\pm$0.20 & 9.11$\pm$0.44 & 7.34$\pm$0.39 & 42.70$\pm$1.59 & 40.42$\pm$2.30 & 4.46$\pm$0.16 & 3.34$\pm$0.13 & 10.19$\pm$0.32 \\
 & AES($k$=10) & 94.22$\pm$0.22 & 6.71$\pm$0.54 & 5.00$\pm$0.44 & 44.31$\pm$2.00 & 39.80$\pm$2.35 & 1.38$\pm$0.15 & 1.94$\pm$0.05 & 8.82$\pm$0.32 \\
 & MCdropout & 94.25$\pm$0.00 & \textbf{5.48$\pm$0.19} & \textbf{3.80$\pm$0.16} & 45.21$\pm$2.19 & \textbf{36.74$\pm$3.06} & 1.45$\pm$0.15 & 1.88$\pm$0.05 & 8.48$\pm$0.13 \\
 & Aleatoric+MC & \textbf{94.33$\pm$0.09} & 6.02$\pm$0.33 & 4.38$\pm$0.30 & \textbf{45.55$\pm$0.87} & 38.72$\pm$1.82 & 1.25$\pm$0.07 & \textbf{1.80$\pm$0.03} & \textbf{8.36$\pm$0.12} \\
 & CRL-softmax & 94.00$\pm$0.12 & 6.02$\pm$0.26 & 4.21$\pm$0.19 & 45.20$\pm$1.15 & 38.81$\pm$1.59 & \textbf{1.23$\pm$0.18} & 1.81$\pm$0.04 & 8.85$\pm$0.20 \\ \hline
\multirow{5}{*}{\begin{tabular}[c]{@{}l@{}}\textbf{CIFAR-10}\\ \textbf{DenseNet}\end{tabular}} & Baseline & 94.87$\pm$0.23 & 5.15$\pm$0.35 & 3.82$\pm$0.30 & 44.21$\pm$2.21 & 36.35$\pm$2.02 & 3.20$\pm$0.20 & 2.23$\pm$0.09 & 8.33$\pm$0.37 \\
 & AES($k$=10) & \textbf{95.00$\pm$0.14} & 5.31$\pm$0.32 & 4.04$\pm$0.26 & 43.29$\pm$1.83 & 37.13$\pm$2.69 & 1.00$\pm$0.10 & 1.66$\pm$0.04 & \textbf{7.65$\pm$0.27} \\
 & MCdropout & 94.69$\pm$0.25 & 5.30$\pm$0.38 & 3.85$\pm$0.28 & 45.64$\pm$2.65 & 36.61$\pm$2.38 & 1.20$\pm$0.09 & 1.73$\pm$0.05 & 7.92$\pm$0.28 \\
 & Aleatoric+MC & 94.73$\pm$0.19 & 5.17$\pm$0.20 & 3.76$\pm$0.14 & \textbf{45.67$\pm$3.18} & \textbf{34.69$\pm$1.03} & 1.25$\pm$0.06 & 1.72$\pm$0.04 & 7.80$\pm$0.16 \\
 & CRL-softmax & 94.71$\pm$0.09 & \textbf{4.92$\pm$0.14} & \textbf{3.49$\pm$0.94} & 45.16$\pm$2.12 & 36.13$\pm$3.35 & \textbf{0.87$\pm$0.07} & \textbf{1.60$\pm$0.02} & 7.84$\pm$0.17 \\ \hline
\multirow{5}{*}{\begin{tabular}[c]{@{}l@{}}\textbf{CIFAR-100}\\ \textbf{VGG-16}\end{tabular}} & Baseline & 73.49$\pm$0.34 & 77.33$\pm$1.15 & 38.61$\pm$0.66 & 68.59$\pm$0.64 & 62.01$\pm$0.39 & 19.81$\pm$0.33 & 17.77$\pm$0.37 & 44.85$\pm$0.51 \\
 & AES($k$=10) & \textbf{74.68$\pm$0.25} & 72.25$\pm$1.13 & 37.09$\pm$0.58 & 67.69$\pm$0.76 & 60.88$\pm$0.92 & 7.42$\pm$0.26 & 10.02$\pm$0.11 & \textbf{35.83$\pm$0.36} \\
 & MCdropout & 73.06$\pm$0.42 & 77.36$\pm$1.15 & 37.85$\pm$0.51 & 67.68$\pm$0.95 & 62.39$\pm$2.16 & 3.37$\pm$0.37 & 10.05$\pm$0.02 & 36.59$\pm$0.29 \\
 & Aleatoric+MC & 73.12$\pm$0.28 & 77.31$\pm$1.00 & 37.43$\pm$0.42 & 67.67$\pm$0.53 & 63.53$\pm$0.81 & \textbf{3.22$\pm$0.19} & \textbf{10.02$\pm$0.04} & 36.63$\pm$0.21 \\
 & CRL-softmax & 74.06$\pm$0.18 & \textbf{71.83$\pm$0.47} & \textbf{34.84$\pm$0.57} & \textbf{69.60$\pm$1.11} & \textbf{59.47$\pm$1.01} & 13.86$\pm$0.27 & 13.10$\pm$0.12 & 39.42$\pm$0.19 \\ \hline
\multirow{5}{*}{\begin{tabular}[c]{@{}l@{}}\textbf{CIFAR-100}\\ \textbf{ResNet110}\end{tabular}} & Baseline & 72.85$\pm$0.30 & 87.24$\pm$1.21 & 46.50$\pm$1.09 & 66.01$\pm$0.43 & 66.03$\pm$1.52 & 16.58$\pm$0.16 & 15.09$\pm$0.14 & 42.83$\pm$0.38 \\
 & AES($k$=10) & 73.65$\pm$0.29 & 79.12$\pm$1.07 & 40.88$\pm$0.49 & 66.72$\pm$0.74 & 63.81$\pm$1.40 & 8.90$\pm$0.15 & 10.67$\pm$0.13 & 37.67$\pm$0.37 \\
 & MCdropout & 74.08$\pm$0.00 & 75.47$\pm$1.07 & 38.53$\pm$1.13 & 66.14$\pm$1.68 & 64.59$\pm$1.46 & 5.35$\pm$0.32 & 10.06$\pm$0.15 & 36.06$\pm$0.38 \\
 & Aleatoric+MC & \textbf{74.50$\pm$0.24} & \textbf{73.26$\pm$0.83} & 37.56$\pm$0.95 & 65.65$\pm$0.91 & 63.53$\pm$1.78 & \textbf{2.68$\pm$0.25} & \textbf{9.24$\pm$0.13} & \textbf{34.96$\pm$0.20} \\
 & CRL-softmax & 74.16$\pm$0.32 & 73.59$\pm$1.39 & \textbf{36.90$\pm$1.08} & \textbf{67.23$\pm$1.13} & \textbf{62.56$\pm$1.26} & 11.52$\pm$0.36 & 10.87$\pm$0.05 & 37.71$\pm$0.44 \\ \hline
\multirow{5}{*}{\begin{tabular}[c]{@{}l@{}}\textbf{CIFAR-100}\\ \textbf{DenseNet}\end{tabular}} & Baseline & 75.39$\pm$0.29 & 71.75$\pm$0.89 & 38.63$\pm$0.72 & 65.18$\pm$1.71 & 63.30$\pm$1.93 & 12.67$\pm$0.25 & 11.54$\pm$0.08 & 37.26$\pm$0.21 \\
 & AES($k$=10) & 76.10$\pm$0.16 & 67.18$\pm$0.37 & 36.04$\pm$0.18 & 64.82$\pm$0.83 & 62.59$\pm$0.69 & 6.78$\pm$0.37 & 9.39$\pm$0.04 & 34.04$\pm$0.14 \\
 & MCdropout & 75.80$\pm$0.36 & 66.92$\pm$1.45 & 34.97$\pm$0.46 & 65.11$\pm$1.10 & 63.27$\pm$1.47 & \textbf{5.59$\pm$0.33} & 9.49$\pm$0.14 & 34.02$\pm$0.38 \\
 & Aleatoric+MC & 75.50$\pm$0.39 & 67.87$\pm$1.55 & 35.05$\pm$0.65 & \textbf{65.92$\pm$1.38} & \textbf{61.69$\pm$1.79} & 6.01$\pm$0.22 & 9.45$\pm$0.13 & 34.25$\pm$0.47 \\
 & CRL-softmax & \textbf{76.82$\pm$0.26} & \textbf{61.77$\pm$1.07} & \textbf{32.57$\pm$0.81} & 65.22$\pm$1.40 & 61.79$\pm$2.20 & 8.59$\pm$0.17 & \textbf{9.11$\pm$0.09} & \textbf{33.39$\pm$0.28} \\ \hline
\multirow{5}{*}{\begin{tabular}[c]{@{}l@{}}\textbf{SVHN}\\ \textbf{VGG-16}\end{tabular}} & Baseline & 96.20$\pm$0.10 & 5.97$\pm$0.28 & 5.24$\pm$0.28 & 41.15$\pm$0.95 & 32.08$\pm$0.56 & 3.15$\pm$0.11 & 2.69$\pm$0.05 & 6.86$\pm$0.17 \\
 & AES($k$=10) & 96.54$\pm$0.09 & 4.59$\pm$0.10 & 3.98$\pm$0.11 & \textbf{43.48$\pm$0.86} & \textbf{27.40$\pm$0.99} & 0.54$\pm$0.09 & 1.34$\pm$0.01 & 5.31$\pm$0.06 \\
 & MCdropout & 96.79$\pm$0.05 & 4.64$\pm$0.34 & 4.12$\pm$0.31 & 41.62$\pm$1.21 & 27.46$\pm$0.95 & \textbf{0.36$\pm$0.02} & \textbf{1.25$\pm$0.03} & \textbf{4.96$\pm$0.11} \\
 & Aleatoric+MC & \textbf{96.80$\pm$0.01} & 4.86$\pm$0.26 & 4.34$\pm$0.26 & 41.14$\pm$0.60 & 27.60$\pm$1.45 & 0.38$\pm$0.07 & 1.26$\pm$0.01 & 4.99$\pm$0.02 \\
 & CRL-softmax & 96.55$\pm$0.07 & \textbf{4.47$\pm$0.10} & \textbf{3.86$\pm$0.08} & 42.82$\pm$1.35 & 29.82$\pm$1.42 & 0.88$\pm$0.12 & 1.52$\pm$0.03 & 5.44$\pm$0.10 \\ \hline
\multirow{5}{*}{\begin{tabular}[c]{@{}l@{}}\textbf{SVHN}\\ \textbf{ResNet110}\end{tabular}} & Baseline & 96.45$\pm$0.06 & 8.02$\pm$0.76 & 7.38$\pm$0.75 & 38.83$\pm$1.79 & 35.78$\pm$1.45 & 2.79$\pm$0.06 & 2.38$\pm$0.04 & 6.25$\pm$0.12 \\
 & AES($k$=10) & 96.77$\pm$0.05 & 4.41$\pm$0.17 & 3.89$\pm$0.16 & \textbf{43.56$\pm$2.51} & \textbf{27.39$\pm$1.34} & \textbf{0.43$\pm$0.11} & 1.26$\pm$0.01 & 4.97$\pm$0.05 \\
 & MCdropout & 97.00$\pm$0.00 & 4.99$\pm$0.35 & 4.53$\pm$0.34 & 39.10$\pm$0.94 & 28.69$\pm$2.22 & 0.65$\pm$0.07 & 1.29$\pm$0.01 & 4.73$\pm$0.13 \\
 & Aleatoric+MC & \textbf{97.01$\pm$0.04} & 5.54$\pm$0.24 & 5.09$\pm$0.23 & 38.71$\pm$1.08 & 31.60$\pm$0.50 & 0.54$\pm$0.05 & \textbf{1.25$\pm$0.01} & \textbf{4.69$\pm$0.05} \\
 & CRL-softmax & 96.81$\pm$0.09 & \textbf{4.25$\pm$0.12} & \textbf{3.74$\pm$0.14} & 43.46$\pm$1.78 & 27.71$\pm$0.56 & 0.85$\pm$0.09 & 1.31$\pm$0.02 & 4.97$\pm$0.12 \\ \hline
\multirow{5}{*}{\begin{tabular}[c]{@{}l@{}}\textbf{SVHN}\\ \textbf{DenseNet}\end{tabular}} & Baseline & 96.40$\pm$0.08 & 7.70$\pm$0.41 & 7.05$\pm$0.39 & 39.43$\pm$0.78 & 34.23$\pm$1.21 & 2.51$\pm$0.07 & 2.10$\pm$0.05 & 6.13$\pm$0.15 \\
 & AES($k$=10) & 96.78$\pm$0.08 & 4.50$\pm$0.16 & 3.98$\pm$0.15 & \textbf{43.43$\pm$1.39} & \textbf{26.16$\pm$1.17} & \textbf{0.41$\pm$0.09} & \textbf{1.24$\pm$0.02} & \textbf{4.96$\pm$0.10} \\
 & MCdropout & 96.82$\pm$0.04 & 5.10$\pm$0.52 & 4.59$\pm$0.51 & 39.57$\pm$2.58 & 31.04$\pm$1.67 & 0.42$\pm$0.06 & 1.29$\pm$0.03 & 4.97$\pm$0.11 \\
 & Aleatoric+MC & \textbf{96.86$\pm$0.14} & 5.68$\pm$1.19 & 5.18$\pm$1.15 & 39.09$\pm$2.28 & 31.43$\pm$3.61 & 0.79$\pm$0.87 & 1.44$\pm$0.35 & 5.05$\pm$0.41 \\
 & CRL-softmax & 96.61$\pm$0.12 & \textbf{4.47$\pm$0.14} & \textbf{3.89$\pm$0.13} & 43.35$\pm$0.81 & 28.35$\pm$1.62 & 0.85$\pm$0.06 & 1.38$\pm$0.04 & 5.26$\pm$0.18 \\ \cline{1-10} 
\end{tabular}}
    \vskip -0.1in
\end{table*}
\endgroup

\section{Experiments}

First, we evaluate our method on the ordinal ranking task with image classification benchmark datasets. Then, the performances on out-of-distribution detection and active learning tasks are presented. The subsequent subsections provide about experimental settings and results of each task. More details on datasets, evaluation metrics, and methods for comparison are available in the supplementary material. Our code is available at \href{https://github.com/daintlab/confidence-aware-learning}{https://github.com/daintlab/confidence-aware-learning}.

\subsection{Ordinal Ranking}
\label{sec-ordinal-ranking}

In this section, we examine how well confidence estimates obtained from a deep classifier trained with CRL is ranked according to the correctness of predictions. This is our primary goal in order to build a classifier being immune to the overconfident prediction issue.

\textbf{Experimental settings.} We evaluate our method on benchmark datasets for image classification: SVHN \citep{netzer2011} and CIFAR-10/100 \citep{krizhevsky2009}. For models to compare, we consider popular deep neural network architectures: VGG-16 \citep{simonyan2014}, PreAct-ResNet110 \citep{he2016} and DenseNet-BC ($k=12, d=100$) \citep{huang2017dense}. All models are trained using SGD with a momentum of 0.9, an initial learning rate of 0.1, and a weight decay of 0.0001 for 300 epochs with the mini-batch size of 128. The learning rate is reduced by a factor of 10 at 150 epochs and 250 epochs. We employ a standard data augmentation scheme, i.e., random horizontal flip and 32$\times$32 random crop after padding with 4 pixels on each side. 

For learning with CRL (CRL model), we set $\lambda$ in Eq.~\eqref{eq-total-loss} to 1.0 without the hyperparameter search process. Note that when estimating confidence from a model trained with CRL, we use the $\kappa$ that is utilized for training. For example, when we set $\kappa$ as the maximum class probability, the confidence function used to evaluate metrics is also the maximum class probability. We compare the performance of CRL model with a standard deep classifier trained with only $\mathcal{L}_{\text{CE}}$ (hereafter referred to as Baseline), MCdropout \citep{gal2016}, Aleatoric+MCdropout \citep{kendall2017} and AES \citep{geifman2018} with 10 and 30 snapshot models. For MCdropout and Aleatoric+MCdropout, entropy on the predicted class probabilities averaged over 50 stochastic predictions is used as uncertainty estimates \citep{kendall2017,corbiere2019}. The maximum of the averaged class probabilities from snapshot models is used to measure confidence for AES \citep{geifman2018}.

\textbf{Evaluation metrics.} We evaluate the quality of confidence estimates in terms of both ordinal ranking and calibration. To measure the ordinal ranking performance, commonly used metrics are employed: the area under the risk-coverage curve (AURC) that is defined to be risk (i.e., error rate) as a function of coverage, \textit{Excess}-AURC (E-AURC) that is a normalized AURC \citep{geifman2018}, the area under the precision-recall curve using errors as the positive class (AUPR-Error) \citep{corbiere2019}, and the false positive rate at 95\% true positive rate (FPR-95\%-TPR). For calibration, we use the expected calibration error (ECE) \citep{naeini2015}, the Brier score \citep{brier1950} and negative log likelihood (NLL).

\begingroup
\renewcommand{\arraystretch}{1.1} 
\begin{table}[t]
    \centering
    \caption{Comparison of ensembles of five classifiers. For each experiment, the best result is shown in boldface. AURC and E-AURC values are multiplied by $10^3$, and NLL are multiplied by $10$ for clarity. All remaining values are percentage.}
    \label{table-clf-ensemble-perf}
    \vskip 0.05in
    \resizebox{0.48\textwidth}{!}{
            \begin{tabular}{ll|ccccc|ccc}
\hline
\multirow{2}{*}{\begin{tabular}[c]{@{}l@{}}\textbf{Dataset}\\\textbf{Model}\end{tabular}} & \multirow{2}{*}{\textbf{Method}} &
\multirow{2}{*}{\begin{tabular}[c]{@{}c@{}}\textbf{\footnotesize{ACC}}\\ \small{($\uparrow$)}\end{tabular}} & \multirow{2}{*}{\begin{tabular}[c]{@{}c@{}}\textbf{\footnotesize{AURC}}\\ \small{($\downarrow$)}\end{tabular}} & \multirow{2}{*}{\begin{tabular}[c]{@{}c@{}}\textbf{\footnotesize{E-AURC}}\\ \small{($\downarrow$)}\end{tabular}} & \multirow{2}{*}{\begin{tabular}[c]{@{}c@{}}\textbf{\footnotesize{AUPR-}}\\ \textbf{\footnotesize{Err}} \footnotesize{($\uparrow$)}\end{tabular}} & \multirow{2}{*}{\begin{tabular}[c]{@{}c@{}}\textbf{\footnotesize{FPR-}}\textbf{\footnotesize{95\%}}\\ \textbf{\small{TPR}} \small{($\downarrow$)}\end{tabular}} & \multirow{2}{*}{\begin{tabular}[c]{@{}c@{}}\textbf{\footnotesize{ECE}}\\ \small{($\downarrow$)}\end{tabular}} & \multirow{2}{*}{\begin{tabular}[c]{@{}c@{}}\textbf{\footnotesize{NLL}}\\ \small{($\downarrow$)}\end{tabular}} & \multirow{2}{*}{\begin{tabular}[c]{@{}c@{}}\textbf{\footnotesize{Brier}}\\ \small{($\downarrow$)}\end{tabular}} \\
 &  &  &  &  &  &  &  &  &  \\ \hline
\multirow{2}{*}{\begin{tabular}[c]{@{}l@{}}\footnotesize{\textbf{CIFAR-10}}\\ \textbf{\small{VGG-16}}\end{tabular}} & \small{Baseline} & 95.02 & 4.45 & 3.19 & \textbf{46.45} & \textbf{33.73} & 1.52 & 1.92 & 7.65 \\
 & \small{CRL-softmax} & \textbf{95.09} & \textbf{4.32} & \textbf{3.09} & 45.27 & 37.88 & \textbf{1.32} & \textbf{1.78} & \textbf{7.51} \\ \hline
\multirow{2}{*}{\begin{tabular}[c]{@{}l@{}}\footnotesize{\textbf{CIFAR-10}}\\ \textbf{\small{ResNet110}}\end{tabular}} & \small{Baseline} & 95.42 & 4.01 & 2.95 & \textbf{44.14} & \textbf{29.03} & 1.12 & 1.63 & 6.86 \\
 & \small{CRL-softmax} & \textbf{95.55} & \textbf{3.72} & \textbf{2.72} & 44.01 & 29.88 & \textbf{0.84} & \textbf{1.50} & \textbf{6.60} \\ \hline
\multirow{2}{*}{\begin{tabular}[c]{@{}l@{}}\footnotesize{\textbf{CIFAR-10}}\\ \textbf{\small{DenseNet}}\end{tabular}} & \small{Baseline} & \textbf{96.03} & \textbf{3.02} & \textbf{2.22} & 44.17 & 30.73 & \textbf{0.79} & 1.29 & \textbf{5.97} \\
 & \small{CRL-softmax} & 95.97 & 3.17 & 2.35 & \textbf{45.25} & \textbf{29.77} & 0.85 & \textbf{1.27} & 5.99 \\ \hline
\multirow{2}{*}{\begin{tabular}[c]{@{}l@{}}\footnotesize{\textbf{CIFAR-100}}\\ \textbf{\small{VGG-16}}\end{tabular}} & \small{Baseline} & 78.34 & 54.53 & 29.16 & 64.99 & 58.44 & 4.07 & 9.53 & 31.05 \\
 & \small{CRL-softmax} & \textbf{78.53} & \textbf{52.53} & \textbf{27.63} & \textbf{66.53} & \textbf{57.89} & \textbf{3.80} & \textbf{9.11} & \textbf{30.47} \\ \hline
\multirow{2}{*}{\begin{tabular}[c]{@{}l@{}}\footnotesize{\textbf{CIFAR-100}}\\ \textbf{\small{ResNet110}}\end{tabular}} & \small{Baseline} & 78.83 & 54.91 & 30.72 & 64.42 & 58.99 & 2.39 & 8.63 & 30.19 \\
 & \small{CRL-softmax} & \textbf{79.08} & \textbf{52.87} & \textbf{29.27} & \textbf{64.88} & \textbf{57.74} & \textbf{2.11} & \textbf{8.06} & \textbf{29.59} \\ \hline
\multirow{2}{*}{\begin{tabular}[c]{@{}l@{}}\footnotesize{\textbf{CIFAR-100}}\\ \textbf{\small{DenseNet}}\end{tabular}} & \small{Baseline} & 80.34 & 47.43 & 26.70 & \textbf{63.83} & \textbf{56.10} & 1.87 & 7.43 & 27.74 \\
 & \small{CRL-softmax} & \textbf{80.85} & \textbf{45.63} & \textbf{25.99} & 61.46 & 57.33 & \textbf{1.79} & \textbf{7.13} & \textbf{27.34} \\ \hline
\multirow{2}{*}{\begin{tabular}[c]{@{}l@{}}\footnotesize{\textbf{SVHN}}\\ \textbf{\small{VGG-16}}\end{tabular}} & \small{Baseline} & 96.91 & 4.48 & 4.00 & \textbf{40.66} & \textbf{28.64} & 1.09 & 1.60 & 4.93 \\
 & \small{CRL-softmax} & \textbf{96.95} & \textbf{4.07} & \textbf{3.60} & 40.52 & 29.25 & \textbf{1.02} & \textbf{1.53} & \textbf{4.92} \\ \hline
\multirow{2}{*}{\begin{tabular}[c]{@{}l@{}}\footnotesize{\textbf{SVHN}}\\ \textbf{\small{ResNet110}}\end{tabular}} & \small{Baseline} & 97.13 & 4.33 & 3.91 & \textbf{42.52} & \textbf{26.30} & 0.92 & 1.38 & 4.47 \\
 & \small{CRL-softmax} & \textbf{97.29} & \textbf{3.80} & \textbf{3.43} & 40.75 & 26.80 & \textbf{0.88} & \textbf{1.23} & \textbf{4.26} \\ \hline
\multirow{2}{*}{\begin{tabular}[c]{@{}l@{}}\footnotesize{\textbf{SVHN}}\\ \textbf{\small{DenseNet}}\end{tabular}} & \small{Baseline} & \textbf{97.24} & 4.93 & 4.55 & 36.49 & 30.54 & \textbf{0.83} & 1.34 & 4.51 \\
 & \small{CRL-softmax} & 97.18 & \textbf{4.10} & \textbf{3.70} & \textbf{43.31} & \textbf{29.05} & 0.87 & \textbf{1.25} & \textbf{4.46} \\ \hline
\end{tabular}
    }
    \vskip -0.1in
\end{table}
\endgroup

\begin{figure}[t]
    \vskip -0.1in
        \centering
        \includegraphics[width=\columnwidth]{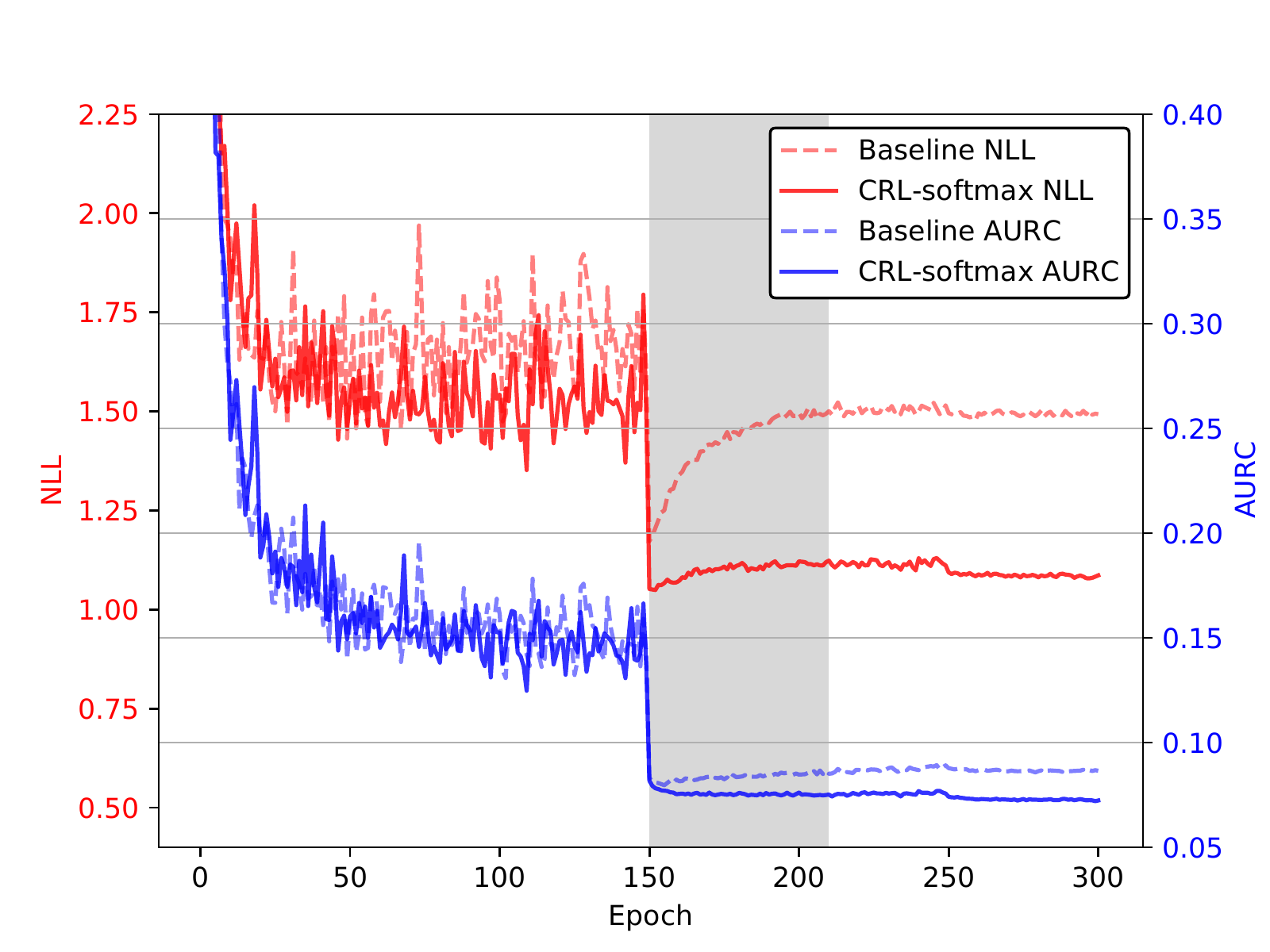}
        \vspace*{-7mm}
        \caption{Comparison of AURC (blue) and NLL (red) curves of Baseline and CRL model with PreAct-ResNet110 on CIFAR-100. Dashed and solid line represents the curves from Baseline and CRL model respectively.}
        \label{fig-aurc-nll}
    \vskip -0.1in
\end{figure}

\begingroup
\renewcommand{\arraystretch}{1.2} 
\begin{table*}[ht]
\centering
\caption{Performances of CRL model on out-of-distribution detection task. The means and standard deviations are computed from five models presented in Section~\ref{sec-ordinal-ranking}. For each comparison, better result is shown in boldface. All values are percentage.
}
\label{table-ood-perf}
    \vskip 0.05in
    \resizebox{\textwidth}{!}{
    \begin{tabular}{llccccc} \toprule
\multicolumn{1}{l}{\begin{tabular}[c]{@{}l@{}}\textbf{In-dist}\\\textbf{Model} \end{tabular}} & \multicolumn{1}{l}{\textbf{Out-of-dist} } & \textbf{FPR-95\%TPR($\downarrow$)}  & \textbf{Detection Err($\downarrow$)}  & \textbf{AUROC($\uparrow$)}  & \textbf{AUPR-In($\uparrow$)}  & \textbf{AUPR-Out($\uparrow$)}  \\ \hline
\multicolumn{1}{c}{} & \multicolumn{1}{c}{} & \multicolumn{5}{c}{\begin{tabular}[c]{@{}c@{}}\textbf{Baseline / CRL}\\\textbf{ Baseline+ODIN / CRL+ODIN }\\\textbf{Baseline+Mahalanobis / CRL+Mahalanobis} \end{tabular}} \\ \cline{3-7}
\multirow{4}{*}{\begin{tabular}[c]{@{}l@{}}\textbf{SVHN} \\\textbf{ResNet110}\\ \end{tabular}} & TinyImageNet & \begin{tabular}[c]{@{}c@{}} 29.65$\pm$2.40 / \textbf{5.89$\pm$0.70\;\;} \\ 27.50$\pm$3.09 / \textbf{2.17$\pm$0.26\;\;} \\ \textbf{0.24$\pm$0.08} / 0.30$\pm$0.12 \end{tabular}& \begin{tabular}[c]{@{}c@{}}12.11$\pm$0.96 / \textbf{5.05$\pm$0.23}\;\; \\ 13.14$\pm$1.29 / \textbf{3.41$\pm$0.23}\;\; \\ \textbf{1.15$\pm$0.16} / 1.39$\pm$0.27 \end{tabular} & \begin{tabular}[c]{@{}c@{}}93.00$\pm$1.06 / \textbf{98.83$\pm$0.13}\\ 92.32$\pm$1.38 / \textbf{99.39$\pm$0.08} \\ \textbf{99.88$\pm$0.03} / 99.82$\pm$0.06 \end{tabular} & \begin{tabular}[c]{@{}c@{}}96.31$\pm$0.91 / \textbf{99.56$\pm$0.04}\\ 95.63$\pm$1.17 / \textbf{99.76$\pm$0.03} \\ \textbf{99.96$\pm$0.01} / 99.93$\pm$0.02 \end{tabular} & \begin{tabular}[c]{@{}c@{}}84.95$\pm$1.41 / \textbf{96.72$\pm$0.50}\\ 85.61$\pm$1.85 / \textbf{98.41$\pm$0.30} \\ \textbf{99.39$\pm$0.19} / 98.99$\pm$0.28 \end{tabular} \\ \cline{3-7}
 & LSUN & \begin{tabular}[c]{@{}c@{}}32.37$\pm$2.78 / \textbf{7.48$\pm$0.91\;\;}\\ 29.57$\pm$3.98 / \textbf{2.92$\pm$0.51\;\;} \\ 0.08$\pm$0.05 / \textbf{0.06$\pm$0.07} \end{tabular} & \begin{tabular}[c]{@{}c@{}} 13.01$\pm$1.17 / \textbf{5.50$\pm$0.20}\;\;\\ 14.06$\pm$1.23 / \textbf{3.88$\pm$0.30}\;\; \\ 0.88$\pm$0.14 / \textbf{0.85$\pm$0.32} \end{tabular} & \begin{tabular}[c]{@{}c@{}} 92.19$\pm$1.39 / \textbf{98.62$\pm$0.17}\\ 91.56$\pm$1.78 / \textbf{99.28$\pm$0.11} \\ \textbf{99.91$\pm$0.03} / 99.89$\pm$0.06 \end{tabular} & \begin{tabular}[c]{@{}c@{}} 95.82$\pm$1.30 / \textbf{99.49$\pm$0.06}\\ 95.19$\pm$1.65 / \textbf{99.72$\pm$0.03} \\ \textbf{99.97$\pm$0.01} / 99.96$\pm$0.02 \end{tabular} & \begin{tabular}[c]{@{}c@{}} 83.48$\pm$1.74 / \textbf{96.14$\pm$0.62} \\ 84.42$\pm$2.38 / \textbf{98.06$\pm$0.42} \\ \textbf{99.45$\pm$0.25} / 99.03$\pm$0.38 \end{tabular} \\ \cline{1-7}
\multirow{5}{*}{\begin{tabular}[c]{@{}l@{}}\textbf{SVHN} \\\textbf{DenseNet}\\\ \end{tabular}} & TinyImageNet & \begin{tabular}[c]{@{}c@{}}26.32$\pm$5.55 / \textbf{7.99$\pm$2.49\;\;} \\ 19.93$\pm$4.43 / \textbf{3.39$\pm$1.34\;\;} \\ 1.44$\pm$1.62 / \textbf{1.03$\pm$1.41} \end{tabular} & \begin{tabular}[c]{@{}c@{}} 11.49$\pm$1.61 / \textbf{5.75$\pm$0.71}\;\;\\ 11.46$\pm$1.80 / \textbf{4.04$\pm$0.80}\;\;\\ 2.42$\pm$1.15 / \textbf{1.86$\pm$0.92} \end{tabular} & \begin{tabular}[c]{@{}c@{}} 93.75$\pm$1.43 / \textbf{98.53$\pm$0.41}\\ 94.06$\pm$1.47 / \textbf{99.17$\pm$0.28} \\ 99.37$\pm$0.91 / \textbf{99.48$\pm$0.79} \end{tabular} & \begin{tabular}[c]{@{}c@{}} 96.62$\pm$0.94 / \textbf{99.43$\pm$0.16}\\ 96.56$\pm$0.94 / \textbf{99.65$\pm$0.12} \\ 99.52$\pm$1.08 / \textbf{99.62$\pm$0.99} \end{tabular} & \begin{tabular}[c]{@{}c@{}} 87.30$\pm$2.74 / \textbf{96.15$\pm$1.16}\\ 90.03$\pm$2.39 / \textbf{98.00$\pm$0.67} \\ 98.45$\pm$1.04 / \textbf{98.48$\pm$0.80} \end{tabular} \\ \cline{3-7}
 & LSUN & \begin{tabular}[c]{@{}c@{}}28.95$\pm$5.80 / \textbf{11.05$\pm$3.09} \\22.22$\pm$4.86 / \textbf{4.63$\pm$1.84\;\;} \\ \textbf{0.41$\pm$0.84} / 0.44$\pm$0.46 \end{tabular} & \begin{tabular}[c]{@{}c@{}} 12.39$\pm$1.84 / \textbf{6.58$\pm$0.74}\;\;\\ 12.35$\pm$1.98 / \textbf{4.68$\pm$0.94}\;\; \\ \textbf{1.23$\pm$0.63} / 1.23$\pm$0.66 \end{tabular} & \begin{tabular}[c]{@{}c@{}} 92.95$\pm$1.76 / \textbf{98.12$\pm$0.49}\\ 93.32$\pm$1.80 / \textbf{98.93$\pm$0.36} \\ 99.73$\pm$0.60 / \textbf{99.75$\pm$0.17} \end{tabular} & \begin{tabular}[c]{@{}c@{}} 96.11$\pm$1.23 / \textbf{99.29$\pm$0.19}\\ 96.15$\pm$1.22 / \textbf{99.56$\pm$0.15} \\ \textbf{99.86$\pm$0.88} / 99.79$\pm$0.13 \end{tabular} & \begin{tabular}[c]{@{}c@{}} 85.93$\pm$3.12 / \textbf{95.06$\pm$1.40}\\ 88.83$\pm$2.73 / \textbf{97.45$\pm$0.89} \\ \textbf{98.97$\pm$0.60} / 98.70$\pm$0.62 \end{tabular} \\ \cline{1-7}
\multirow{5}{*}{\begin{tabular}[c]{@{}l@{}}\textbf{CIFAR-10} \\\textbf{ResNet110}\\\ \end{tabular}} & TinyImageNet & \begin{tabular}[c]{@{}c@{}} 66.09$\pm$2.86 / \textbf{53.17$\pm$5.60}\\ 49.33$\pm$4.19 / \textbf{43.08$\pm$5.15} \\ \textbf{8.46$\pm$2.12} / 9.44$\pm$2.25 \end{tabular} & \begin{tabular}[c]{@{}c@{}} 22.59$\pm$1.81 / \textbf{22.06$\pm$2.35}\\ 22.08$\pm$2.28 / \textbf{17.69$\pm$2.02} \\ \textbf{6.39$\pm$0.87} / 7.02$\pm$0.91 \end{tabular} & \begin{tabular}[c]{@{}c@{}} 82.59$\pm$2.91 / \textbf{86.25$\pm$2.76}\\ 84.31$\pm$3.22 / \textbf{90.40$\pm$1.91} \\ \textbf{98.34$\pm$0.41} / 97.92$\pm$0.41 \end{tabular} & \begin{tabular}[c]{@{}c@{}} 79.63$\pm$5.39 / \textbf{86.56$\pm$3.27}\\ 80.73$\pm$5.07 / \textbf{90.77$\pm$2.13} \\ \textbf{98.40$\pm$0.37} / 97.85$\pm$0.35 \end{tabular} & \begin{tabular}[c]{@{}c@{}} 82.07$\pm$2.00 / \textbf{85.61$\pm$2.50}\\ 86.01$\pm$2.30 / \textbf{90.03$\pm$1.77} \\ \textbf{98.22$\pm$0.49} / 98.02$\pm$0.44 \end{tabular} \\ \cline{3-7}
 & LSUN & \begin{tabular}[c]{@{}c@{}} 57.65$\pm$2.89 / \textbf{44.53$\pm$6.57}\\ 34.72$\pm$5.75 / \textbf{32.10$\pm$5.29} \\ 6.33$\pm$2.54 / \textbf{5.52$\pm$1.39} \end{tabular} & \begin{tabular}[c]{@{}c@{}} \textbf{17.78$\pm$1.21} / 17.89$\pm$1.87\\ 16.29$\pm$1.76 / \textbf{13.50$\pm$1.64} \\ 5.51$\pm$1.25 / \textbf{5.16$\pm$0.76} \end{tabular} & \begin{tabular}[c]{@{}c@{}} 88.25$\pm$1.54 / \textbf{90.46$\pm$1.93}\\ 90.63$\pm$1.97 / \textbf{93.90$\pm$1.23} \\ 98.66$\pm$0.51 / \textbf{98.71$\pm$0.29} \end{tabular} & \begin{tabular}[c]{@{}c@{}} 87.73$\pm$2.59 / \textbf{91.37$\pm$1.95}\\ 88.98$\pm$2.79 / \textbf{94.48$\pm$1.21} \\ \textbf{98.79$\pm$0.48} / 98.76$\pm$0.25 \end{tabular} & \begin{tabular}[c]{@{}c@{}} 87.06$\pm$1.29 / \textbf{89.60$\pm$1.97}\\ 91.47$\pm$1.66 / \textbf{93.30$\pm$1.25} \\ 98.49$\pm$0.61 / \textbf{98.62$\pm$0.31} \end{tabular} \\ \cline{1-7}
\multirow{4}{*}{\begin{tabular}[c]{@{}l@{}}\textbf{CIFAR-10} \\\textbf{DenseNet}\\ \end{tabular}} & TinyImageNet & \begin{tabular}[c]{@{}c@{}} 45.81$\pm$3.95 / \textbf{29.87$\pm$4.09}\\ 10.73$\pm$6.24 / \textbf{10.41$\pm$3.09} \\ 6.99$\pm$1.13 / \textbf{6.28$\pm$3.18} \end{tabular} & \begin{tabular}[c]{@{}c@{}} 13.15$\pm$1.41 / \textbf{12.99$\pm$1.03}\\ 7.09$\pm$2.02 / \textbf{6.89$\pm$1.10} \\ 5.92$\pm$0.58 / \textbf{5.61$\pm$1.54} \end{tabular} & \begin{tabular}[c]{@{}c@{}}93.25$\pm$1.04 / \textbf{94.50$\pm$0.84}\\ 97.86$\pm$1.09 / \textbf{97.97$\pm$0.56} \\ 98.37$\pm$0.50 / \textbf{98.52$\pm$1.17} \end{tabular} & \begin{tabular}[c]{@{}c@{}} 94.53$\pm$0.94 / \textbf{95.17$\pm$0.71}\\ 97.90$\pm$1.04 / \textbf{98.16$\pm$0.48} \\ \textbf{98.22$\pm$1.29} / 98.09$\pm$2.07 \end{tabular} & \begin{tabular}[c]{@{}c@{}} 91.82$\pm$1.24 / \textbf{93.87$\pm$1.03}\\ \textbf{97.84$\pm$1.13} / 97.78$\pm$0.65 \\ 98.49$\pm$0.38 / \textbf{98.57$\pm$0.82} \end{tabular} \\ \cline{3-7}
 & LSUN & \begin{tabular}[c]{@{}c@{}} 36.31$\pm$3.64 / \textbf{21.22$\pm$2.73}\\ \textbf{4.32$\pm$2.55} / 5.29$\pm$1.53 \\ 5.27$\pm$1.15 / \textbf{3.86$\pm$2.15} \end{tabular} & \begin{tabular}[c]{@{}c@{}} 10.60$\pm$0.88 / \textbf{10.59$\pm$0.55}\\ \textbf{4.46$\pm$1.15} / 5.03$\pm$0.71 \\ 5.08$\pm$0.57 / \textbf{4.26$\pm$1.11} \end{tabular} & \begin{tabular}[c]{@{}c@{}} 95.18$\pm$0.64 / \textbf{96.34$\pm$0.44}\\ \textbf{99.04$\pm$0.46} / 98.81$\pm$0.29 \\ 98.73$\pm$0.50 / \textbf{98.89$\pm$0.66} \end{tabular} & \begin{tabular}[c]{@{}c@{}} 96.16$\pm$0.50 / \textbf{96.80$\pm$0.35}\\ \textbf{99.10$\pm$0.41} / 98.92$\pm$0.24 \\ \textbf{98.68$\pm$1.56} / 98.67$\pm$1.07 \end{tabular} & \begin{tabular}[c]{@{}c@{}} 94.14$\pm$0.86 / \textbf{95.94$\pm$0.57}\\ \textbf{98.99$\pm$0.50} / 98.70$\pm$0.35 \\ 98.71$\pm$0.36 / \textbf{98.91$\pm$0.50} \end{tabular} \\ \bottomrule
\end{tabular}
}
\vskip -0.1in
\end{table*}
\endgroup

\textbf{Results.\footnote{Due to the space limitation, we present a subset of results. Complete results can be found in the supplementary material.}} Comparative results are summarized in Table~\ref{table-clf-perf}. CRL-softmax in the table means CRL model using the maximum class probability as a confidence estimator. From the results, we observe that a standard deep classifier trained with CRL improves both classification accuracy and the quality of confidence estimates of Baseline. For example, in case of DenseNet on CIFAR-100, CRL-softmax has 1.43\% higher accuracy than Baseline and shows greatly improved confidence estimates evaluated over all performance metrics. It implies that CRL is an effective regularizer encouraging a classifier to produce good probabilistic predictions. Surprisingly, we observe that CRL model outputs comparable or better confidence estimates compared to MCdropout, Aleatoric+MCdropout and AES which require multiple stochastic predictions or snapshot models. For instance, CRL model outperforms the competing methods in 7 cases among all 9 experiments in terms of AURC. The results demonstrate that training with CRL is very effective to build a reliable and strong baseline being comparable to such methods. 

We also examine whether Deep Ensembles \citep{lakshminarayanan2017} benefits from CRL. Table~\ref{table-clf-ensemble-perf} presents the comparison results of ensembles based on five Baseline and five CRL models. For these experiments, we set $\lambda$ to $0.5$ for CRL models since we empirically found that the ensemble of CRL models with $\lambda=1$ does not improve Baseline ensemble on CIFAR-10 except other datasets (refer to Table~\ref{stable-clf-ensemble-weight1} in the supplementary material). We infer that it is because CRL acts as a strong regularizer so the trained CRL models with a large $\lambda$ from random initial points lose their diversity. Thus, we use smaller $\lambda$ to address this diversity issue. With $\lambda=0.5$, it is observed that CRL also improves Deep Ensembles. One notable point from the results is that although CRL is designed to learn better confidence estimates in terms of ordinal ranking, it is also beneficial to calibration performance. 

To further understand the effect of CRL, NLL and AURC curves on CIFAR-100 test set are shown in Figure~\ref{fig-aurc-nll}. NLL curves from Baseline and CRL model show that CRL effectively regularizes a classifier. Also, in Baseline model, we can observe a natural trend that overfitting to NLL leads to poor ordinal ranking as can be seen in the shaded area. Remarkably, training with CRL, however, further improves the ranking performance even when the model slightly overfits to NLL. This observation supports the regularization effect on training a classifier with CRL.

\subsection{Out-of-Distribution Detection (OOD)}
OOD detection is the problem of identifying inputs that come from the distribution (i.e., out-of-distribution) sufficiently different from the training distribution (i.e., in-distribution). Through the experiments, we demonstrate that a classifier trained with CRL separate well in- and out-of-distribution samples.

\textbf{Experimental settings.} Following \citet{devries2018}, we use two in-distribution datasets: SVHN and CIFAR-10. For the out-of-distribution datasets, we use TinyImageNet\footnote{\href{https://tiny-imagenet.herokuapp.com/}{https://tiny-imagenet.herokuapp.com/}}, LSUN \citep{yu2015}, and iSUN \citep{xu2015}. Also, we utilize five Baseline and CRL-softmax models that are trained previously for Section~\ref{sec-ordinal-ranking}.

First, we compare the OOD detection performance of Baseline models with CRL-softmax models. Then, we investigate whether ODIN \citep{liang2018} and the Mahalanobis detector \citep{lee2018_mahal} combined with CRL models can improve the detection performance further. ODIN and Mahalanobis are post-processing methods that boost the OOD detection performance of a pre-trained classifier significantly, which have the hyperparameters: a temperature $T$ for ODIN, and a perturbation magnitude $\epsilon$ for both ODIN and Mahalanobis. To determine the hyperparameter values, we employ the procedure described in \citet{lee2018_mahal}.\footnote{For the experiment, we used the code publicly available at \href{https://github.com/pokaxpoka/deep\_Mahalanobis\_detector}{https://github.com/pokaxpoka/deep\_Mahalanobis\_detector}.}

\textbf{Evaluation metrics.} We employ five metrics commonly used for the task \citep{hendrycks2017,devries2018}: FPR-at-95\%-TPR, detection error that measures the minimum classification error over all possible thresholds, the area under the receiver operating characteristic curve (AUROC), the area under the precision-recall curve using in-distribution samples as the positives (AUPR-In), and AUPR using out-of-distribution samples as the positives (AUPR-Out).

\textbf{Results.} Comparing the performance of Baseline and CRL-softmax models, CRL models perform better in most cases with a large margin as shown in Table~\ref{table-ood-perf}.\footnote{We omit the results where the OOD dataset is iSUN since it shows similar results with LSUN. Please refer to Table~\ref{stable-ood-perf} in the supplementary material.} This means that the CRL model provides good confidence estimates that distinguish OOD samples from in-distribution ones much more easily. 
We also observe that ODIN indeed becomes a more reliable detector when combined with CRL model. It outperforms ODIN with Baseline in all experiments with the exception of DenseNet with CIFAR-10 on LSUN, the OOD dataset. Interestingly, CRL model by itself performs even better than ODIN where SVHN dataset is the in-distribution dataset. For example, in case of FPR-95\%-TPR for TinyImageNet OOD dataset with DenseNet, the values from CRL-softmax (i.e., 7.99) is significantly lower than those from Baseline ODIN (i.e., 19.93), and we find similar results for the remaining metrics. The Mahalanobis detector is already a strong OOD detector on the datasets we consider, but it also slightly benefits from CRL models although the performance improvements are marginal compared to ODIN. Note that the conventional experimental setting for OOD detection is disadvantageous to CRL models since our models are trained to produce low confidence even for in-distribution samples if they are misclassified. Nevertheless, our experimental results show that deep classifiers trained with CRL perform well on the OOD detection task under that setting.

\subsection{Active Learning}
The key hypothesis of active learning lies that we can build a good predictive model with less labeled samples if a model knows which samples should be labeled to improve predictive performance. Thus, the goal of active learning is to achieve greater accuracy with fewer training labels \citep{settles2009}. 

\begin{figure}[!t]
  \vskip 0.1in
      \centering
      \subfigure[CIFAR-10]{
      \includegraphics[width=0.90\columnwidth]{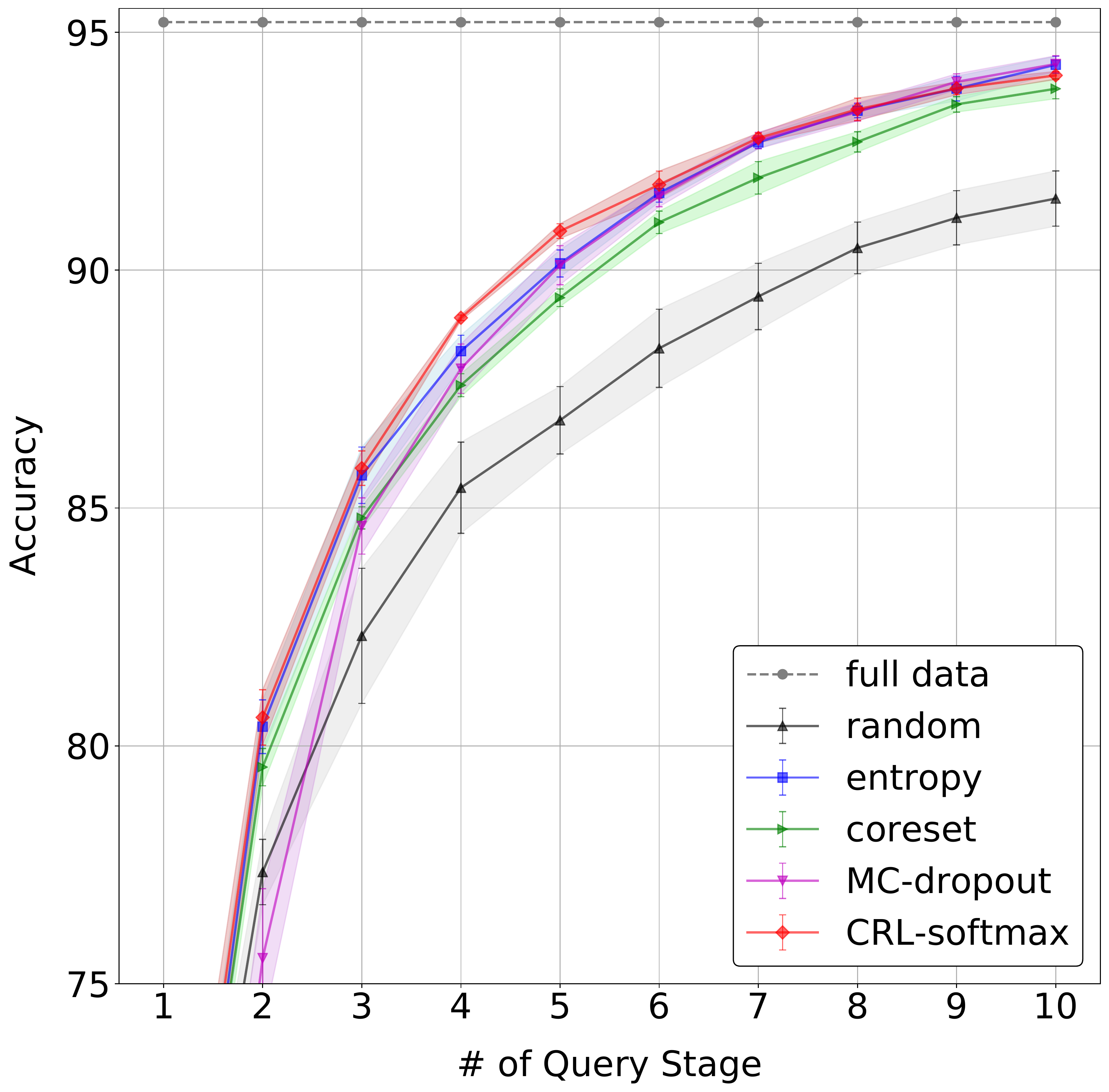}
      \label{fig-al-cifar10}
      }
      \subfigure[CIFAR-100]{
      \includegraphics[width=0.90\columnwidth]{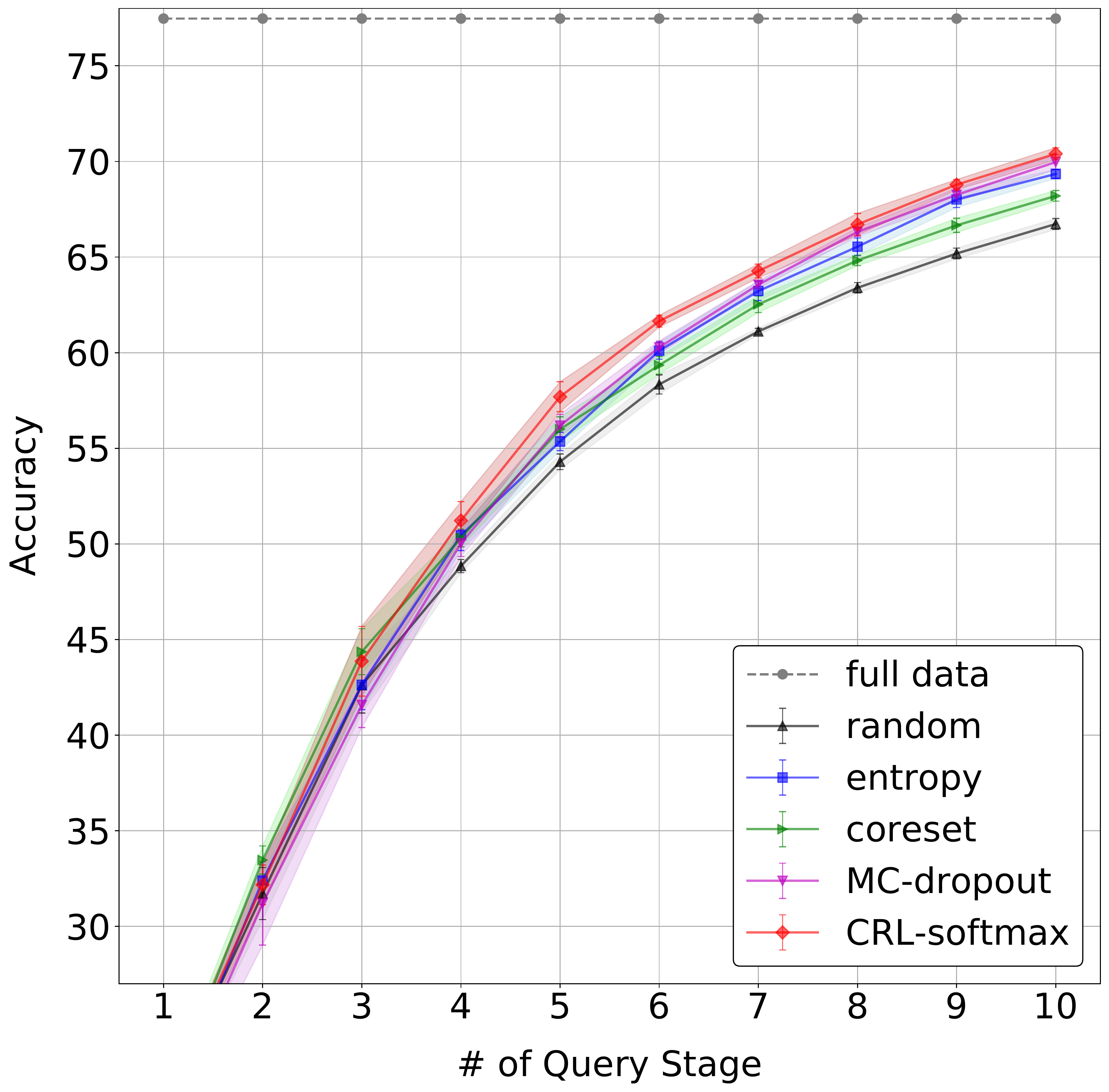}
      \label{fig-al-cifar100}
      }
      \caption{Active learning performance on (a) CIFAR-10 and (b) CIFAR-100 with various sampling methods. Curves are averages over five runs, and shaded areas denote $\pm$ one standard deviation. 
      }
  \vskip -0.1in
  \end{figure}

\textbf{Experimental settings.} We evaluate the active learning performance of CRL model with ResNet18 architecture\footnote{\href{https://github.com/kuangliu/pytorch-cifar}{https://github.com/kuangliu/pytorch-cifar}} by using CIFAR-10 and CIFAR-100 datasets. In this experiment, we train the model during 200 epochs, and decay the learning rate with a factor of 10 at 120 and 160 epochs. Other hyperparameters involved in training are same as in Section~\ref{sec-ordinal-ranking}. For a comparison, we consider a CRL-softmax model associated with the least confidence-based sampling, MCdropout with entropy-based sampling, and Baseline with core-set sampling \citep{sener2018} designed specifically for active learning to query representative samples. As other baselines commonly employed in active learning, we also use Baseline with random sampling and entropy-based sampling.  

For this task, we follow a typical process to evaluate the performance of sampling strategy for active learning \citep{sener2018,yoo2019}. Given a unlabeled dataset $\mathcal{D}^0_U$ (i.e., the whole 50,000 images without labels), the labeled dataset $\mathcal{D}^1_L$ at the first stage  consists of 2,000 samples that are randomly sampled without replacement from $\mathcal{D}^0_U$. With $\mathcal{D}^1_L$, we train an initial model.
According to uncertainty estimates from the model, additional 2,000 samples are added to the labeled dataset for the next stage, and this $\mathcal{D}^2_L$ is used to update the current model. We proceed a total of 10 stages for a single trial. To rigorously compare the performances, we repeat this trial five times.

\textbf{Results.} Figure~\ref{fig-al-cifar10} shows the performance improvement over the stages on CIFAR-10. Obviously, Baseline with random sampling is inferior to other methods. CRL-softmax with the least confidence sampling shows better performance than the competing methods for most of stages. At the second stage, CRL-softmax has 80.6\% of accuracy while entropy-based, core-set, and MCdropout have 80.40\%, 79.55\%, and 75.53\% accuracy respectively. Our method also shows the highest performance compared to others at the 6-th stage. It reaches to 91.8\% accuracy at this stage while entropy-based, core-set, MCdropout sampling methods show 0.2\%, 0.8\%, and 0.25\% lower accuracy than our model.

The performance curves on CIFAR-100 can be found in Figure~\ref{fig-al-cifar100}. Since CIFAR-100 is a more challenging dataset than CIFAR-10, it is comparatively hard to learn with small labeled dataset at early stages. Nevertheless, CRL model selects most of the informative samples that should be labeled, thereby showing better performance for all stages after the 4-th one. Finally, CRL model is the only one that achieves over 70\% (i.e., 70.4\%) accuracy. It shows the 0.43\% accuracy gap with MCdropout, the second-best performing model. 

Apart from CRL model, it is observed that Baseline with entropy-based sampling performs quite well on both datasets even if it is one of the most simple approaches, as similarly reported in \citet{yoo2019}. It should be mentioned that the core-set sampling is a query strategy to enhance active learning performance, and MCdropout method needs multiple stochastic forward paths to estimate uncertainty. Through the experimental results, we demonstrate that good confidence estimates naturally obtained from the CRL model are indeed effective for active learning.

\section{Conclusion}

In this paper, we introduce a simple but effective regularization method that can be employed for training deep neural networks to alleviate the well-known overconfident prediction issue. Our method is motivated by the observation regarding the correct prediction events during training with the SGD-based optimizer. Based on that, the proposed regularization method is implemented by the ranking loss CRL, which greatly improves confidence ranking performance of deep classifiers. We have demonstrated that deep neural networks trained with CRL produce well-ranked confidence estimates that are particularly important to the tasks related to what the model does not know such as OOD detection and active learning. Although we apply the proposed method to image classification tasks in the experiments, it can be extended to other classification tasks in natural language processing. It would be also interesting to investigate other properties of the proposed method such as its robustness to adversarial samples. 

\section*{Acknowledgements}

This research was supported by Basic Science Research Program through the National Research Foundation of Korea(NRF) funded by the Ministry of Education(NRF-2018R1D1A1A02086017).

\bibliography{paper}
\bibliographystyle{icml2020}

\clearpage

\icmltitlerunning{Supplementary Materials: Confidence-Aware Learning for Deep Neural Networks}

\twocolumn[
\icmltitle{\textit{Supplementary Materials}:\\Confidence-Aware Learning for Deep Neural Networks}

\vskip 0.3in
]

\renewcommand{\thesection}{S\arabic{section}}
\setcounter{section}{0}
\renewcommand\thefigure{S\arabic{figure}}
\setcounter{figure}{0}
\renewcommand\thetable{S\arabic{table}}
\setcounter{table}{0}
\renewcommand{\theHtable}{Supplement.\thetable}
\renewcommand{\theHfigure}{Supplement.\thefigure}

\section{Experimental Details: Ordinal Ranking}
\label{ssec-ordinal-ranking}

\subsection{Evaluation Metrics}

\textbf{AURC \& E-AURC}
AURC measures the area under the curve drawn by plotting the risk according to coverage. The coverage indicates the ratio of samples whose confidence estimates are higher than some confidence threshold, and the risk, also known as the selective risk \citep{geifman2017}, is an error rate computed by using those samples. A low value of AURC implies that correct and incorrect predictions can be well-separable by confidence estimates associated with samples.

Inherently, AURC is affected by the predictive performance of a model. To have a unitless performance measure that can be applied across models, \citet{geifman2018} introduce a normalized AURC, named Excess-AURC (E-AURC). E-AURC can be computed by subtracting the optimal AURC, the lowest possible value for a given model, from the empirical AURC. For a detailed description, please refer to \citet{geifman2018}.

\textbf{AUPR-Error}
AUPR measures the area under the precision-recall curve. The precision-recall curve is a graph showing the precision~=~TP/(TP+FP) against recall~=~TP/(TP+FN), where TP, FP, and FN denote true positives, false positives, and false negatives, respectively. The AUPR-ERROR represents the area under precision-recall curve where misclassified samples (i.e., errors) are used as positives. This is used as the primary metric to evaluate the failure prediction performance in \citet{corbiere2019}.

\textbf{FPR-at-95\%-TPR}
FPR-at-95\%-TPR measures the false positive rate~(FPR)~=~FP/(FP+TN) when the true positive rate~(TPR)~=~TP/(TP+FN) is 95\%, where TP, TN, FP, and FN denotes true positives, true negatives, false positives, and false negatives, respectively. It can be interpreted as the probability that an example predicted incorrectly is misclassified as a correct prediction when TPR is equal to 95\%.

\textbf{ECE}
Expected calibration error (ECE) \citep{naeini2015} is a metric that approximates the difference in expectation between accuracy and confidence. As an approximation, ECE partitions the probability interval into a fixed number of bins. Then, each bin $B_m$ has an interval $(\frac{m-1}{M},\frac{m}{M}],m=1,...,M$ where $M$ is the number of bins. With these bins, ECE can be computed as
\begin{equation*}
\label{eq-ECE}
    \textnormal{ECE}=\sum\limits_{m=1}^{M}{\frac{|B_m|}{n}\left|\textnormal{acc}(B_m)-\textnormal{conf}(B_m)\right|}
\end{equation*}
where $n$ is the total number of samples, $\textnormal{acc}(B_m)$ denotes the accuracy computed from samples in $B_m$, and $\textnormal{conf}(B_m)$ is the average confidence scores of samples in $B_m$. 

\textbf{NLL}
Negative log likelihood~(NLL) is a standard measure for evaluating the quality of predictive probability, which is computed as
\begin{equation*}
    \textnormal{NLL} = -\sum_{i=1}^{n} {\log {P\left( y=y_i \middle|\mathbf{x}_i, \mathbf{w} \right)}}.
\end{equation*}

\textbf{Brier Score}
Brier score \citep{brier1950} can be interpreted as the average mean squared error between the predicted probability and one-hot encoded label. It can be computed as
\begin{equation*}
    \textnormal{Brier}=\frac{1}{n}\sum\limits_{i=1}^{n}\sum\limits_{k=1}^{K}
    (P\left( y=k \middle|\mathbf{x}_i, \mathbf{w} \right)-t_{k})^2
\end{equation*}
where $t_k=1$ if $k=y_i$, and 0 otherwise.

\subsection{Experimental Settings}

\textbf{Datasets} 
CIFAR-10 and CIFAR-100 are the datasets for a multi-class image classification task. They consist of 50K training images and 10K test images of size $32\times32$ with 10 and 100 classes, respectively. The Street View House Numbers (SVHN) dataset \citep{netzer2011} contains 73,257 training images and 26,032 test images of size $32\times32$ with 10 classes of digits. 

\textbf{MCdropout} 
VGG-16 for MCdropout is the one used in \citet{geifman2018}.\footnote{\href{https://github.com/geifmany/uncertainty\_ICLR}{https://github.com/geifmany/uncertainty\_ICLR}} 
Specifically, a dropout layer with a dropout rate $p=0.3$ is added after the first convolutional layer, and dropout layers with $p=0.4$ are applied to other convolutional layers except ones followed by a max pooling layer. For fully connected layers, dropout with $p=0.5$ is used. 
PreAct-ResNet110 for MCdropout comes from \citet{zhang2019}. Dropout layers with $p=0.2$ are applied to all convolutional layer, and a dropout layer with $p=0.1$ is added before the last fully connected layer. Note that this architecture from \citet{zhang2019} was determined through the validation process.
DenseNet-BC already has dropout layers and we set the dropout rate of them to 0.2 as used in the original paper \citep{huang2017}.
In the experiments, we compute 50 stochastic predictions and the entropy on the average predicted probabilities is used as an uncertainty estimate.

\textbf{Aleatoric+MCdropout}
To consider aleatoric uncertainty, a Gaussian distribution whose mean is a model's prediction is placed over the logit space as proposed in \citet{kendall2017}. The models for Aleatoric+MCdropout are the same as used for MCdropout except that an additional output layer is attached to produce the variance of the Gaussian distribution. With this Gaussian distribution, 50 logit vectors are sampled and averaged to compute a cross-entropy loss during training. Like MCdropout, we use 50 stochastic predictions and the entropy is used to estimate uncertainty.

\textbf{AES} 
Average early stopping (AES) is a snapshot ensemble approach motivated by the observation that easy samples are learned earlier during training while hard samples are not. To leverage this for confidence estimation, AES method provides the average confidence estimates from the ensemble of model snapshots. \citet{geifman2018} suggests an ensemble with $k$ models at epochs $i\in F$ where $F$ is a set of $k$ evenly spaced integers between $0.4T$ and $T$. Here, $T$ denotes the total number of epochs. In the experiments, we consider $k=10$ and $k=30$.

\subsection{Results}

Table~\ref{stable-clf-perf-cifar10}, \ref{stable-clf-perf-cifar100} and \ref{stable-clf-perf-svhn} shows the complete experimental results to evaluate ordinal ranking performance on CIFAR-10, CIFAR-100 and SVHN, respectively. For CRL models, we consider the maximum class probability (CRL-softmax), negative entropy (CRL-entropy), and margin (CRL-margin) as a confidence function, respectively. Regardless of the confidence function, it is observed that CRL improves the quality of confidence estimates. Compared to other methods that require multiple predictions, CRL models consistently yield comparable or better performance. 

\begin{figure*}[h]
    \centering
    \vspace*{-2mm}
    \subfigure[CIFAR-10]{
    \includegraphics[width=1.1\columnwidth]{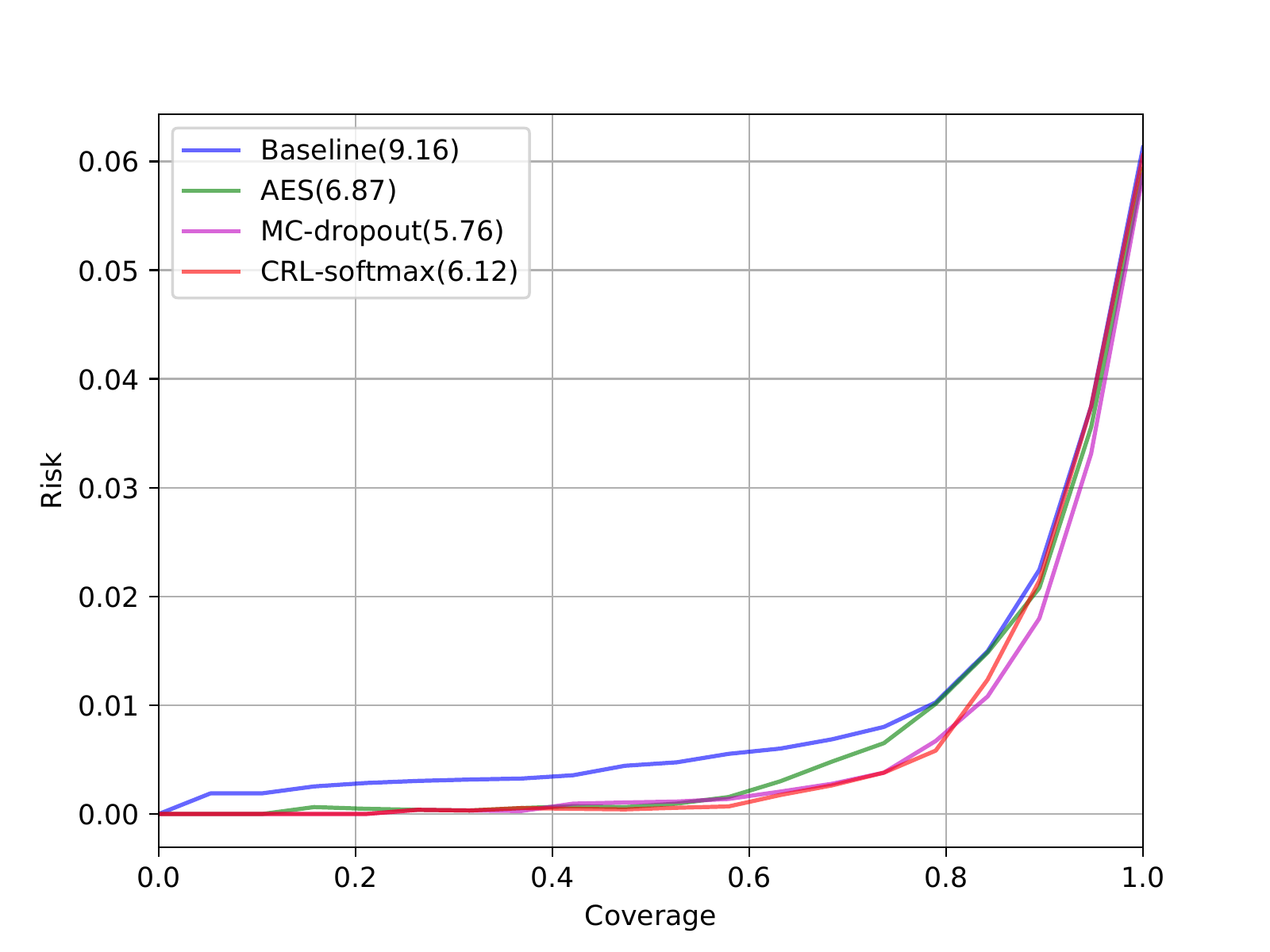}
    \label{sfig-cifar10-rc}
    }
    \vspace*{-2mm}
    \subfigure[CIFAR-100]{
    
    \includegraphics[width=1.1\columnwidth]{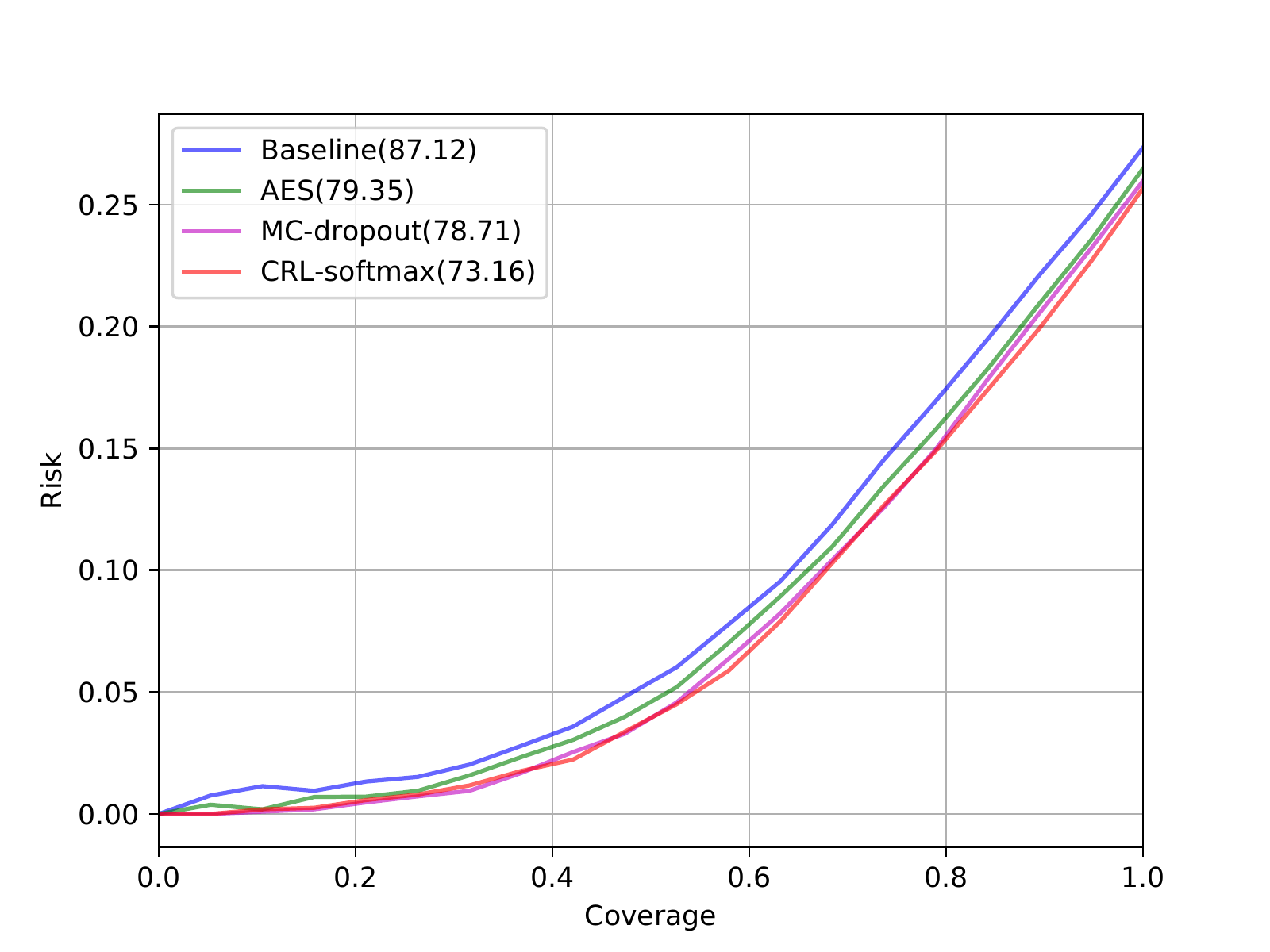}
    \label{sfig-cifar100-rc}
    } 
    \subfigure[SVHN]{
    \includegraphics[width=1.1\columnwidth]{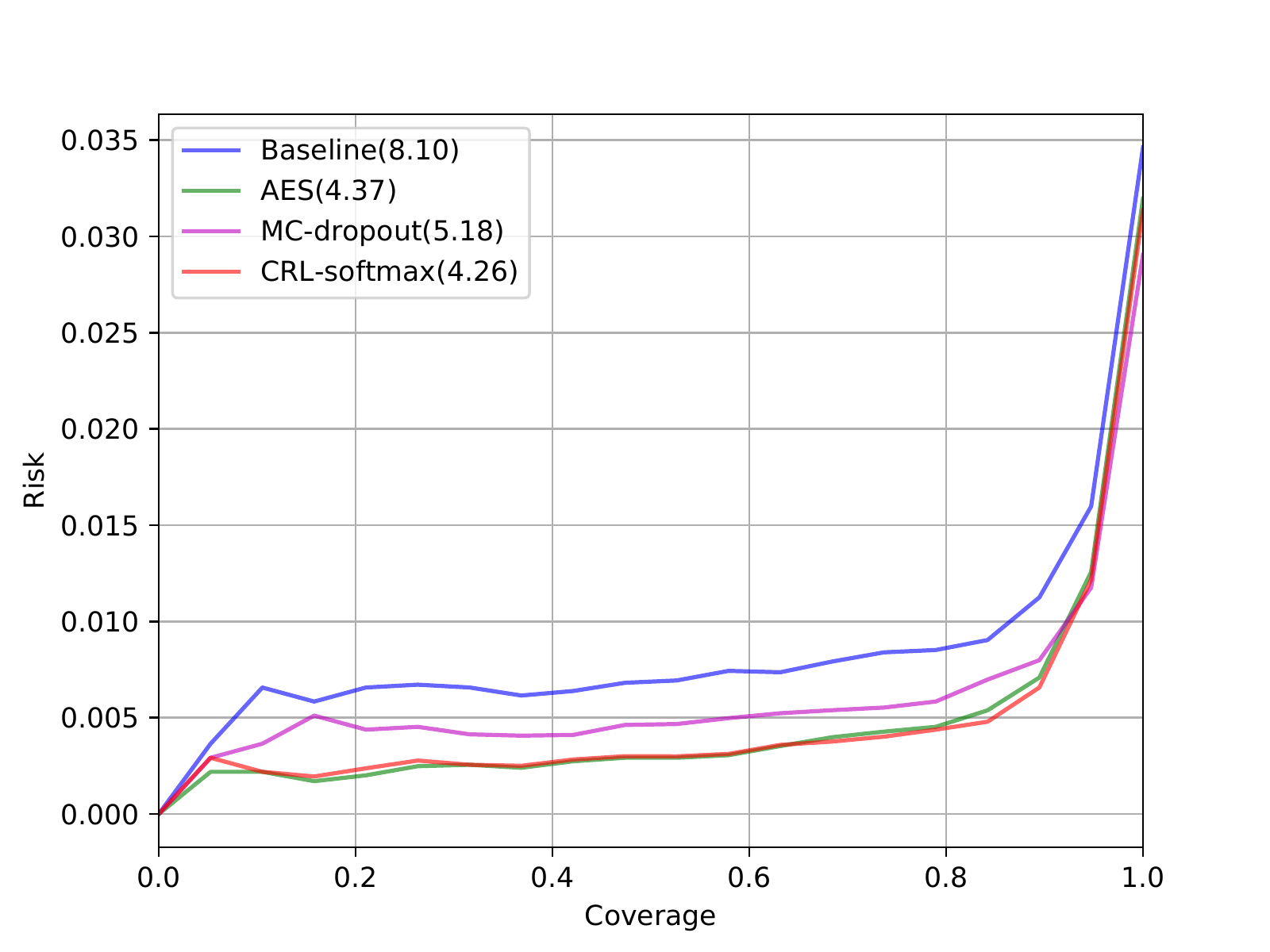}
    \label{sfig-svhn-rc}
    }

\caption{Risk-coverage curves from PreAct-ResNet110 on (a) CIFAR-10, (b) CIFAR-100, and (c) SVHN.}
\label{sfig-rc}
\end{figure*}

Figure~\ref{sfig-rc} shows the risk-coverage~(RC) curve plots from PreAct-ResNet110 on CIFAR-10/100 and SVHN datasets. A score in parentheses is the AURC value associated with each model. For this figure, the model that shows the median performance among five repeated runs is selected.

Tables~\ref{stable-clf-ensemble-weight0.5} and \ref{stable-clf-ensemble-weight1} show ordinal ranking performance of CRL ensembles with $\lambda=0.5$ and $\lambda=1.0$, respectively.

\section{Experimental Details: Out-of-Distribution Detection}
\label{ssec-ood}

\subsection{Evaluation metrics} 

\textbf{Detection Error} Detection error measures the minimum possible error rate over all possible thresholds when separating in- and out-of-distribution samples.

\textbf{AUROC} The area under the receiver operating characteristic curve (AUROC) measures the area under the curve drawn by plotting the true positive rate against the false positive rate.

\textbf{AUPR-In \& AUPR-Out} AUPR measures the area under the precision-recall curve. AUPR-In and AUPR-Out use in- and out-of-distribution samples as positives, respectively. 

\subsection{Experimental Settings}

\textbf{Datasets} 
The TinyImageNet is a subset of ImageNet dataset that contains 10,000 test images with 200 classes. The LSUN dataset consists of 10,000 images of 10 different scenes \citep{yu2015}. The iSUN dataset is a subset of LSUN images and consists of 8,925 images. These datasets are used as out-of-distribution datasets, and all images are resized to $32\times32$.

\textbf{ODIN} ODIN (Out-of-DIstribution detector for Neural networks) \citep{liang2018} is a simple and effective post-processing method for out-of-distribution detection. ODIN consists of two steps: temperature scaling and adding small perturbations to inputs. Through a manipulation of temperature constant $T$, the softmax scores of in- and out-of-distribution images can be distinguishable by pusing them further apart from each other. In addition, an input is preprocessed by adding small perturbations to decrease the softmax score. The perturbations can be computed as the gradient of loss with respect to the input, and they are added to the input with a multiplicative constant $\epsilon$. To find the hyperparameters $T$ and $\epsilon$, a small hold-out set from out-of-distribution dataset was used following to the procedure in the original paper.    

\textbf{Mahalanobis} 
\citet{lee2018_mahal} proposed the Mahalanobis distance-based confidence score to identify out-of-distribution samples from the finding that the trained features of deep neural networks follow the class-conditional Gaussian distribution. To further enhance the detection performance, it adds small perturbations $\epsilon$ to an input similar to ODIN, and combines the confidence scores from all layers in a deep neural network. Concretely, the scores are computed by weighted averaging and these weights are determined by training a logistic regression model using a validation dataset. The optimal value of $\epsilon$ was chosen via validation process as described in the original paper.

\subsection{Results}
Table~\ref{stable-ood-perf} shows full out-of-distribution detection results including those from iSUN dataset. Since iSUN is a subset of LSUN, the detection performances on iSUN are similar to those on LSUN.

\section{Experimental Details: Active Learning}
\label{ssec-al}

\subsection{Experimental Settings}

Since query strategies for active learning are based on uncertainty, there exists a risk that samples selected to be labeled are overlapped, i.e., they might have redundant information. To avoid this issue, we select the samples from a random subset of the unlabeled pool $\mathcal{D}_U^S$ at $S$-th stage. We set the size of subset to 10,000. \citet{beluch2018} and \citet{yoo2019} are also used this simple scheme to address the redundancy issue.

The proposed method requires counting correct prediction events of all training samples. Hence, incremental learning with newly labeled samples cannot be applied to CRL models. For a fair comparison, we initialize all models including comparison targets at the beginning of every stage, i.e., all models are trained from scratch with their labeled dataset. To control the unexpected effect of random initialization, all models share the same random seed at each stage.

\textbf{Query Strategy} 
We consider the following query strategies (i.e., sampling methods) for comparison: random sampling, entropy-based sampling, core-set sampling \citep{sener2018}, and entropy-based sampling with MCdropout. Random sampling is selecting samples to be labeled randomly. Entropy-based sampling selects samples whose entropy of predicted class probability is high. Entropy-based sampling with MCdropout differs from just entropy-based sampling in that it measures entropy on the average predicted class probabilities obtained by 50 stochastic predictions. Core-set sampling focuses on the representativeness of samples, which can be implemented by K-Center-Greedy algorithm. Following to \citet{sener2018}, we use the $l_2$ distance between activations of the last fully connected layer to measure the diversity of samples.

\subsection{Results}
Table~\ref{stable-al-perf} shows the classification accuracy values of sampling strategies at each active learning stage.

\begin{table*}[h]
    \centering
    \caption{Comparison of the quality of confidence estimates on CIFAR-10. The means and standard deviations over five runs are reported. $\downarrow$ and $\uparrow$ indicate that lower and higher values are better respectively. AURC and E-AURC values are multiplied by $10^3$, and NLL are multiplied by $10$ for clarity. All remaining values are percentage. \textcolor{red}{\textbf{Red}} and \textcolor{blue}{\textbf{blue}} represent the best performance among single models and the methods requiring multiple predictions, respectively.}
    \label{stable-clf-perf-cifar10}
    \vskip 0.1in
    \resizebox{0.95\textwidth}{!}{
    \begin{tabular}{ll|ccccc|ccc}
\hline
\multirow{2}{*}{\textbf{\begin{tabular}[c]{@{}l@{}}\textbf{Dataset}\\ \textbf{Model}\end{tabular}}} & \multirow{2}{*}{\textbf{\textbf{Method}}} & \multirow{2}{*}{\textbf{\begin{tabular}[c]{@{}c@{}}\textbf{\small{ACC}}\\ \small{($\uparrow$)}\end{tabular}}} & \multirow{2}{*}{\textbf{\begin{tabular}[c]{@{}c@{}}\textbf{\small{AURC}}\\ \small{($\downarrow$)}\end{tabular}}} & \multirow{2}{*}{\textbf{\begin{tabular}[c]{@{}c@{}}\textbf{\small{E-AURC}}\\ \small{($\downarrow$)}\end{tabular}}} & \multirow{2}{*}{\textbf{\begin{tabular}[c]{@{}c@{}}\textbf{\small{AUPR-}}\\ \textbf{\small{Err}} \footnotesize{($\uparrow$)}\end{tabular}}} & \multirow{2}{*}{\textbf{\begin{tabular}[c]{@{}c@{}}\textbf{\small{FPR-95\%}}\\ \textbf{\small{TPR}} \small{($\downarrow$)}\end{tabular}}} & \multirow{2}{*}{\begin{tabular}[c]{@{}c@{}}\textbf{\small{ECE}}\\ \small{($\downarrow$)}\end{tabular}} & \multirow{2}{*}{\begin{tabular}[c]{@{}c@{}}\textbf{\small{NLL}}\\ \small{($\downarrow$)}\end{tabular}} & \multirow{2}{*}{\begin{tabular}[c]{@{}c@{}}\textbf{\small{Brier}}\\ \small{($\downarrow$)}\end{tabular}} \\
 &  &  &  &  &  &  &  &  &  \\ \hline
\multirow{8}{*}{\begin{tabular}[c]{@{}l@{}}\textbf{CIFAR-10}\\ \textbf{VGG-16}\end{tabular}} & \multicolumn{1}{l|}{Baseline} & 93.74$\pm$0.14 & 7.10$\pm$0.31 & 5.10$\pm$0.26 & 44.19$\pm$0.34 & \multicolumn{1}{c|}{41.43$\pm$0.38} & 5.20$\pm$0.11 & 3.79$\pm$0.11 & 11.30$\pm$0.21 \\
 & \multicolumn{1}{l|}{CRL-entropy} & 93.84$\pm$0.12 & \color{red}\textbf{6.77$\pm$0.16} & 4.83$\pm$0.16 & 46.16$\pm$2.87 & \multicolumn{1}{c|}{41.35$\pm$3.03} & 2.47$\pm$0.19 & 2.47$\pm$0.03 & 9.99$\pm$0.09 \\
 & \multicolumn{1}{l|}{CRL-softmax} & 93.82$\pm$0.18 & 6.78$\pm$0.18 & \color{red}\textbf{4.83$\pm$0.08} & \color{red}\textbf{46.79$\pm$1.75} & \multicolumn{1}{c|}{\color{red}\textbf{40.21$\pm$2.18}} & \color{red}\textbf{1.24$\pm$0.20} & \color{red}\textbf{2.09$\pm$0.04} & \color{red}\textbf{9.33$\pm$0.21} \\
 & \multicolumn{1}{l|}{CRL-margin} & \color{red}\textbf{93.88$\pm$0.12} & 7.13$\pm$0.23 & 5.21$\pm$0.16 & 43.26$\pm$1.79 & \multicolumn{1}{c|}{44.20$\pm$0.94} & 1.55$\pm$0.13 & 2.73$\pm$0.07 & 9.81$\pm$0.16 \\ \cline{2-10} 
 & \multicolumn{1}{l|}{MCdropout} & 93.78$\pm$0.27 & 6.72$\pm$0.28 & 4.72$\pm$0.19 & 45.08$\pm$2.14 & \multicolumn{1}{c|}{41.52$\pm$2.83} & 1.11$\pm$0.19 & 1.93$\pm$0.05 & 9.34$\pm$0.39 \\
 & \multicolumn{1}{l|}{Aleatoric+MC} & 93.91$\pm$0.13 & 6.57$\pm$0.29 & 4.68$\pm$0.22 & 44.67$\pm$1.76 & \multicolumn{1}{c|}{41.68$\pm$1.86} & \color{blue}\textbf{0.86$\pm$0.12} & \color{blue}\textbf{1.89$\pm$0.05} & \color{blue}\textbf{9.08$\pm$0.24} \\
 & \multicolumn{1}{l|}{AES(k=10)} & \color{blue}\textbf{93.97$\pm$0.12} & 7.15$\pm$0.25 & 5.30$\pm$0.25 & 44.47$\pm$1.00 & \multicolumn{1}{c|}{41.01$\pm$1.75} & 1.61$\pm$0.27 & 2.06$\pm$0.04 & 9.26$\pm$0.15 \\
 & \multicolumn{1}{l|}{AES(k=30)} & 93.96$\pm$0.17 & \color{blue}\textbf{6.50$\pm$0.10} & \color{blue}\textbf{4.64$\pm$0.09} & \color{blue}\textbf{45.36$\pm$3.02} & \multicolumn{1}{c|}{\color{blue}\textbf{38.60$\pm$1.51}} & 1.82$\pm$0.25 & 1.95$\pm$0.03 & 9.23$\pm$0.15 \\ \hline
\multirow{8}{*}{\begin{tabular}[c]{@{}l@{}}\textbf{CIFAR-10}\\ \textbf{ResNet110}\end{tabular}} & \multicolumn{1}{l|}{Baseline} & 94.11$\pm$0.20 & 9.11$\pm$0.44 & 7.34$\pm$0.39 & 42.70$\pm$1.59 & \multicolumn{1}{c|}{40.42$\pm$2.30} & 4.46$\pm$0.16 & 3.34$\pm$0.13 & 10.19$\pm$0.32 \\
 & \multicolumn{1}{l|}{CRL-entropy} & \color{red}\textbf{94.24$\pm$0.11} & \color{red}\textbf{6.01$\pm$0.18} & 4.33$\pm$0.13 & 43.15$\pm$0.43 & \multicolumn{1}{c|}{41.65$\pm$2.66} & \color{red}\textbf{0.79$\pm$0.12} & 1.97$\pm$0.02 & \color{red}\textbf{8.74$\pm$0.12} \\
 & \multicolumn{1}{l|}{CRL-softmax} & 94.00$\pm$0.12 & 6.02$\pm$0.26 & \color{red}\textbf{4.21$\pm$0.19} & 45.20$\pm$1.15 & \multicolumn{1}{c|}{\color{red}\textbf{38.81$\pm$1.59}} & 1.23$\pm$0.18 & \color{red}\textbf{1.81$\pm$0.04} & 8.85$\pm$0.20 \\
 & \multicolumn{1}{l|}{CRL-margin} & 93.83$\pm$0.10 & 6.28$\pm$0.13 & 4.34$\pm$0.07 & \color{red}\textbf{45.46$\pm$1.07} & \multicolumn{1}{c|}{39.92$\pm$1.27} & 1.12$\pm$0.16 & 1.87$\pm$0.01 & 9.07$\pm$0.09 \\ \cline{2-10} 
 & \multicolumn{1}{l|}{MCdropout} & 94.25$\pm$0.00 & \color{blue}\textbf{5.48$\pm$0.19} & \color{blue}\textbf{3.80$\pm$0.16} & 45.21$\pm$2.19 & \multicolumn{1}{c|}{36.74$\pm$3.06} & 1.45$\pm$0.15 & 1.88$\pm$0.05 & 8.48$\pm$0.13 \\
 & \multicolumn{1}{l|}{Aleatoric+MC} & \color{blue}\textbf{94.33$\pm$0.09} & 6.02$\pm$0.33 & 4.38$\pm$0.30 & 45.55$\pm$0.87 & \multicolumn{1}{c|}{38.72$\pm$1.82} & \color{blue}\textbf{1.25$\pm$0.07} & \color{blue}\textbf{1.80$\pm$0.03} & \color{blue}\textbf{8.36$\pm$0.12} \\
 & \multicolumn{1}{l|}{AES(k=10)} & 94.22$\pm$0.22 & 6.71$\pm$0.54 & 5.00$\pm$0.44 & 44.31$\pm$2.00 & \multicolumn{1}{c|}{39.80$\pm$2.35} & 1.38$\pm$0.15 & 1.94$\pm$0.05 & 8.82$\pm$0.32 \\
 & \multicolumn{1}{l|}{AES(k=30)} & 94.20$\pm$0.23 & 5.80$\pm$0.28 & 4.09$\pm$0.25 & \color{blue}\textbf{47.15$\pm$1.93} & \multicolumn{1}{c|}{\color{blue}\textbf{36.37$\pm$2.85}} & 1.61$\pm$0.20 & 1.82$\pm$0.04 & 8.69$\pm$0.29 \\ \hline
\multirow{8}{*}{\begin{tabular}[c]{@{}l@{}}\textbf{CIFAR-10}\\ \textbf{DenseNet}\end{tabular}} & \multicolumn{1}{l|}{Baseline} & 94.87$\pm$0.23 & 5.15$\pm$0.35 & 3.82$\pm$0.30 & 44.21$\pm$2.21 & \multicolumn{1}{c|}{36.35$\pm$2.02} & 3.20$\pm$0.20 & 2.23$\pm$0.09 & 8.33$\pm$0.37 \\
 & \multicolumn{1}{l|}{CRL-entropy} & \color{red}\textbf{94.98$\pm$0.15} & 4.95$\pm$0.30 & 3.67$\pm$0.26 & 40.67$\pm$1.50 & \multicolumn{1}{c|}{42.12$\pm$2.06} & \color{red}\textbf{0.69$\pm$0.15} & 1.66$\pm$0.03 & \color{red}\textbf{7.67$\pm$0.19} \\
 & \multicolumn{1}{l|}{CRL-softmax} & 94.71$\pm$0.09 & \color{red}\textbf{4.92$\pm$0.14} & \color{red}\textbf{3.49$\pm$0.94} & 45.16$\pm$2.12 & \multicolumn{1}{c|}{\color{red}\textbf{36.13$\pm$3.35}} & 0.87$\pm$0.07 & \color{red}\textbf{1.60$\pm$0.02} & 7.84$\pm$0.17 \\
 & \multicolumn{1}{l|}{CRL-margin} & 94.42$\pm$0.19 & 5.26$\pm$0.23 & 3.68$\pm$0.18 & \color{red}\textbf{45.36$\pm$3.22} & \multicolumn{1}{c|}{36.67$\pm$2.19} & 0.95$\pm$0.11 & 1.65$\pm$0.03 & 8.17$\pm$0.18 \\ \cline{2-10} 
 & \multicolumn{1}{l|}{MCdropout} & 94.69$\pm$0.25 & 5.30$\pm$0.38 & 3.85$\pm$0.28 & 45.64$\pm$2.65 & \multicolumn{1}{c|}{36.61$\pm$2.38} & 1.20$\pm$0.09 & 1.73$\pm$0.05 & 7.92$\pm$0.28 \\
 & \multicolumn{1}{l|}{Aleatoric+MC} & 94.73$\pm$0.19 & 5.17$\pm$0.20 & 3.76$\pm$0.14 & \color{blue}\textbf{45.67$\pm$3.18} & \multicolumn{1}{c|}{34.69$\pm$1.03} & 1.25$\pm$0.06 & 1.72$\pm$0.04 & 7.80$\pm$0.16 \\
 & \multicolumn{1}{l|}{AES(k=10)} & \color{blue}\textbf{95.00$\pm$0.14} & 5.31$\pm$0.32 & 4.04$\pm$0.26 & 43.29$\pm$1.83 & \multicolumn{1}{c|}{37.13$\pm$2.69} & \color{blue}\textbf{1.00$\pm$0.10} & 1.66$\pm$0.04 & 7.65$\pm$0.27 \\
 & AES(k=30) & 94.99$\pm$0.18 & \color{blue}\textbf{4.70$\pm$0.20} & \color{blue}\textbf{3.43$\pm$0.15} & 45.39$\pm$2.02 & \color{blue}\textbf{34.37$\pm$1.70} & 1.18$\pm$0.09 & \color{blue}\textbf{1.58$\pm$0.04} & \color{blue}\textbf{7.57$\pm$0.26} \\ \cline{1-10} 
\end{tabular}}%
\end{table*}

\begin{table*}[h]
    \centering
    \caption{Comparison of the quality of confidence estimates on CIFAR-100. The means and standard deviations over five runs are reported. $\downarrow$ and $\uparrow$ indicate that lower and higher values are better respectively. AURC and E-AURC values are multiplied by $10^3$, and NLL are multiplied by $10$ for clarity. All remaining values are percentage. \textcolor{red}{\textbf{Red}} and \textcolor{blue}{\textbf{blue}} represent the best performance among single models and the methods requiring multiple predictions, respectively.}
    \label{stable-clf-perf-cifar100}
    \vskip 0.1in
    \resizebox{0.95\textwidth}{!}{
    \begin{tabular}{ll|ccccc|ccc}
\hline
\multirow{2}{*}{\textbf{\begin{tabular}[c]{@{}l@{}}\textbf{Dataset}\\ \textbf{Model}\end{tabular}}} & \multirow{2}{*}{\textbf{\textbf{Method}}} & \multirow{2}{*}{\textbf{\begin{tabular}[c]{@{}c@{}}\textbf{\small{ACC}}\\ \small{($\uparrow$)}\end{tabular}}} & \multirow{2}{*}{\textbf{\begin{tabular}[c]{@{}c@{}}\textbf{\small{AURC}}\\ \small{($\downarrow$)}\end{tabular}}} & \multirow{2}{*}{\textbf{\begin{tabular}[c]{@{}c@{}}\textbf{\small{E-AURC}}\\ \small{($\downarrow$)}\end{tabular}}} & \multirow{2}{*}{\textbf{\begin{tabular}[c]{@{}c@{}}\textbf{\small{AUPR-}}\\ \textbf{\small{Err}} \footnotesize{($\uparrow$)}\end{tabular}}} & \multirow{2}{*}{\textbf{\begin{tabular}[c]{@{}c@{}}\textbf{\small{FPR-95\%}}\\ \textbf{\small{TPR}} \small{($\downarrow$)}\end{tabular}}} & \multirow{2}{*}{\begin{tabular}[c]{@{}c@{}}\textbf{\small{ECE}}\\ \small{($\downarrow$)}\end{tabular}} & \multirow{2}{*}{\begin{tabular}[c]{@{}c@{}}\textbf{\small{NLL}}\\ \small{($\downarrow$)}\end{tabular}} & \multirow{2}{*}{\begin{tabular}[c]{@{}c@{}}\textbf{\small{Brier}}\\ \small{($\downarrow$)}\end{tabular}} \\
 &  &  &  &  &  &  &  &  &  \\ \hline
\multirow{8}{*}{\begin{tabular}[c]{@{}l@{}}\textbf{CIFAR-100}\\ \textbf{VGG-16}\end{tabular}} & Baseline & 73.49$\pm$0.34 & 77.33$\pm$1.15 & 38.61$\pm$0.66 & 68.59$\pm$0.64 & 62.01$\pm$0.39 & 19.81$\pm$0.33 & 17.77$\pm$0.37 & 44.85$\pm$0.51 \\
 & CRL-entropy & \color{red}\textbf{74.71$\pm$0.19} & \color{red}\textbf{70.19$\pm$1.53} & 35.11$\pm$1.13 & 68.70$\pm$1.08 & \color{red}\textbf{59.15$\pm$2.19} & \color{red}\textbf{11.62$\pm$0.32} & \color{red}\textbf{12.42$\pm$0.10} & \color{red}\textbf{38.16$\pm$0.39} \\
 & CRL-softmax & 74.06$\pm$0.18 & 71.83$\pm$0.47 & \color{red}\textbf{34.84$\pm$0.57} & \color{red}\textbf{69.60$\pm$1.11} & 59.47$\pm$1.01 & 13.86$\pm$0.27 & 13.10$\pm$0.12 & 39.42$\pm$0.19 \\
 & CRL-margin & 74.06$\pm$0.27 & 75.91$\pm$0.76 & 38.93$\pm$0.74 & 67.59$\pm$1.04 & 59.74$\pm$1.62 & 12.16$\pm$0.24 & 13.67$\pm$0.16 & 38.79$\pm$0.36 \\ \cline{2-10} 
 & MCdropout & 73.06$\pm$0.42 & 77.36$\pm$1.15 & 37.85$\pm$0.51 & 67.68$\pm$0.95 & 62.39$\pm$2.16 & 3.37$\pm$0.37 & 10.05$\pm$0.02 & 36.59$\pm$0.29 \\
 & Aleatoric+MC & 73.12$\pm$0.28 & 77.31$\pm$1.00 & 37.43$\pm$0.42 & 67.67$\pm$0.53 & 63.53$\pm$0.81 & \color{blue}\textbf{3.22$\pm$0.19} & 10.02$\pm$0.04 & 36.63$\pm$0.21 \\
 & AES(k=10) & 74.68$\pm$0.25 & 72.25$\pm$1.13 & 37.09$\pm$0.58 & 67.69$\pm$0.76 & \color{blue}\textbf{60.88$\pm$0.92} & 7.42$\pm$0.26 & 10.02$\pm$0.11 & 35.83$\pm$0.36 \\
 & AES(k=30) & \color{blue}\textbf{74.78$\pm$0.30} & \color{blue}\textbf{68.99$\pm$1.24} & \color{blue}\textbf{34.13$\pm$0.74} & \color{blue}\textbf{67.72$\pm$0.95} & 61.20$\pm$1.40 & 7.85$\pm$0.30 & \color{blue}\textbf{9.64$\pm$0.19} & \color{blue}\textbf{35.64$\pm$0.38} \\ \hline
\multirow{8}{*}{\begin{tabular}[c]{@{}l@{}}\textbf{CIFAR-100}\\ \textbf{ResNet110}\end{tabular}} & Baseline & 72.85$\pm$0.30 & 87.24$\pm$1.21 & 46.50$\pm$1.09 & 66.01$\pm$0.43 & 66.03$\pm$1.52 & 16.58$\pm$0.16 & 15.09$\pm$0.14 & 42.83$\pm$0.38 \\
 & CRL-entropy & 73.73$\pm$0.38 & 75.77$\pm$1.81 & 37.78$\pm$1.01 & \color{red}\textbf{67.62$\pm$1.32} & \color{red}\textbf{61.83$\pm$1.46} & \color{red}\textbf{10.37$\pm$0.40} & 11.23$\pm$0.15 & 38.03$\pm$0.53 \\
 & CRL-softmax & 74.16$\pm$0.32 & 73.59$\pm$1.39 & \color{red}\textbf{36.90$\pm$1.08} & 67.23$\pm$1.13 & 62.56$\pm$1.26 & 11.52$\pm$0.36 & 10.87$\pm$0.05 & 37.71$\pm$0.44 \\
 & CRL-margin & \color{red}\textbf{74.66$\pm$0.13} & \color{red}\textbf{73.26$\pm$0.30} & 38.04$\pm$0.56 & 63.27$\pm$0.59 & 66.64$\pm$1.33 & 10.77$\pm$0.21 & \color{red}\textbf{10.50$\pm$0.12} & \color{red}\textbf{36.93$\pm$0.23} \\ \cline{2-10} 
 & MCdropout & 74.08$\pm$0.00 & 75.47$\pm$1.07 & 38.53$\pm$1.13 & 66.14$\pm$1.68 & 64.59$\pm$1.46 & 5.35$\pm$0.32 & 10.06$\pm$0.15 & 36.06$\pm$0.38 \\
 & Aleatoric+MC & \color{blue}\textbf{74.50$\pm$0.24} & \color{blue}\textbf{73.26$\pm$0.83} & \color{blue}\textbf{37.56$\pm$0.95} & 65.65$\pm$0.91 & \color{blue}\textbf{63.53$\pm$1.78} & \color{blue}\textbf{2.68$\pm$0.25} & \color{blue}\textbf{9.24$\pm$0.13} & \color{blue}\textbf{34.96$\pm$0.20} \\
 & AES(k=10) & 73.65$\pm$0.29 & 79.12$\pm$1.07 & 40.88$\pm$0.49 & 66.72$\pm$0.74 & 63.81$\pm$1.40 & 8.90$\pm$0.15 & 10.67$\pm$0.13 & 37.67$\pm$0.37 \\
 & AES(k=30) & 73.67$\pm$0.32 & 76.69$\pm$1.32 & 38.52$\pm$0.96 & \color{blue}\textbf{67.13$\pm$0.76} & 64.23$\pm$0.95 & 9.33$\pm$0.20 & 10.17$\pm$0.11 & 37.61$\pm$0.39 \\ \hline
\multirow{8}{*}{\begin{tabular}[c]{@{}l@{}}\textbf{CIFAR-100}\\ \textbf{DenseNet}\end{tabular}} & Baseline & 75.39$\pm$0.29 & 71.75$\pm$0.89 & 38.63$\pm$0.72 & 65.18$\pm$1.71 & 63.30$\pm$1.93 & 12.67$\pm$0.25 & 11.54$\pm$0.08 & 37.26$\pm$0.21 \\
 & CRL-entropy & 76.24$\pm$0.28 & 64.33$\pm$1.19 & 33.56$\pm$0.54 & \color{red}\textbf{65.36$\pm$0.28} & \color{red}\textbf{61.36$\pm$0.92} & \color{red}\textbf{8.02$\pm$0.39} & 9.60$\pm$0.09 & 34.04$\pm$0.44 \\
 & CRL-softmax & 76.82$\pm$0.26 & 61.77$\pm$1.07 & \color{red}\textbf{32.57$\pm$0.81} & 65.22$\pm$1.40 & 61.79$\pm$2.20 & 8.59$\pm$0.17 & 9.11$\pm$0.09 & 33.39$\pm$0.28 \\
 & CRL-margin & \color{red}\textbf{77.09$\pm$0.18} & \color{red}\textbf{61.51$\pm$0.99} & 33.00$\pm$0.65 & 61.73$\pm$0.64 & 64.23$\pm$1.35 & 8.42$\pm$0.17 & \color{red}\textbf{8.97$\pm$0.10} & \color{red}\textbf{33.06$\pm$0.28} \\ \cline{2-10} 
 & MCdropout & 75.80$\pm$0.36 & 66.92$\pm$1.45 & 34.97$\pm$0.46 & 65.11$\pm$1.10 & 63.27$\pm$1.47 & \color{blue}\textbf{5.59$\pm$0.33} & 9.42$\pm$0.14 & 34.02$\pm$0.38 \\
 & Aleatoric+MC & 75.50$\pm$0.39 & 67.87$\pm$1.55 & 35.05$\pm$0.65 & 65.92$\pm$1.38 & \color{blue}\textbf{61.69$\pm$1.79} & 6.01$\pm$0.22 & 9.45$\pm$0.13 & 34.25$\pm$0.47 \\
 & AES(k=10) & \color{blue}\textbf{76.10$\pm$0.16} & 67.18$\pm$0.37 & 36.04$\pm$0.18 & 64.82$\pm$0.83 & 62.59$\pm$0.69 & 6.78$\pm$0.37 & 9.39$\pm$0.04 & 34.04$\pm$0.14 \\
 & AES(k=30) & 76.05$\pm$0.12 & \color{blue}\textbf{65.22$\pm$0.73} & \color{blue}\textbf{33.95$\pm$0.68} & \color{blue}\textbf{65.94$\pm$0.84} & 62.17$\pm$0.54 & 7.38$\pm$0.22 & \color{blue}\textbf{9.04$\pm$0.04} & \color{blue}\textbf{33.96$\pm$0.16} \\ \cline{1-10} 
\end{tabular}
    }
\end{table*}

\begin{table*}[h]
    \centering
    \caption{Comparison of the quality of confidence estimates on SVHN. The means and standard deviations over five runs are reported. $\downarrow$ and $\uparrow$ indicate that lower and higher values are better respectively. AURC and E-AURC values are multiplied by $10^3$, and NLL are multiplied by $10$ for clarity. All remaining values are percentage. \textcolor{red}{\textbf{Red}} and \textcolor{blue}{\textbf{blue}} represent the best performance among single models and the methods requiring multiple predictions, respectively.}
    \label{stable-clf-perf-svhn}
    \vskip 0.1in
    \resizebox{0.95\textwidth}{!}{
    \begin{tabular}{ll|ccccc|ccc}
\hline
\multirow{2}{*}{\textbf{\begin{tabular}[c]{@{}l@{}}\textbf{Dataset}\\ \textbf{Model}\end{tabular}}} & \multirow{2}{*}{\textbf{\textbf{Method}}} & \multirow{2}{*}{\textbf{\begin{tabular}[c]{@{}c@{}}\textbf{\small{ACC}}\\ \small{($\uparrow$)}\end{tabular}}} & \multirow{2}{*}{\textbf{\begin{tabular}[c]{@{}c@{}}\textbf{\small{AURC}}\\ \small{($\downarrow$)}\end{tabular}}} & \multirow{2}{*}{\textbf{\begin{tabular}[c]{@{}c@{}}\textbf{\small{E-AURC}}\\ \small{($\downarrow$)}\end{tabular}}} & \textbf{\textbf{\small{AUPR-}}} & \textbf{\textbf{\small{FPR-95\%}}} & \multirow{2}{*}{\begin{tabular}[c]{@{}c@{}}\textbf{\small{ECE}}\\ \small{($\downarrow$)}\end{tabular}} & \multirow{2}{*}{\begin{tabular}[c]{@{}c@{}}\textbf{\small{NLL}}\\ \small{($\downarrow$)}\end{tabular}} & \multirow{2}{*}{\begin{tabular}[c]{@{}c@{}}\textbf{\small{Brier}}\\ \small{($\downarrow$)}\end{tabular}} \\
 &  &  &  &  & \textbf{\small{~~~~Err.}} \footnotesize{($\uparrow$)} & \textbf{\small{TPR}} \small{($\downarrow$)} &  &  &  \\ \hline
\multirow{8}{*}{\begin{tabular}[]{@{}l@{}}\textbf{SVHN}\\\textbf{VGG-16}\end{tabular}} & Baseline & 96.20$\pm$0.10 & 5.97$\pm$0.28 & 5.24$\pm$0.28 & 41.15$\pm$0.95 & 32.08$\pm$0.56 & 3.15$\pm$0.11 & 2.69$\pm$0.05 & 6.86$\pm$0.17 \\
 & CRL-entropy & 96.55$\pm$0.10 & \color{red}\textbf{4.31$\pm$0.10} & \color{red}\textbf{3.72$\pm$0.10} & \color{red}\textbf{44.39$\pm$2.87} & \color{red}\textbf{28.34$\pm$1.07} & 1.15$\pm$0.10 & 1.55$\pm$0.03 & 5.54$\pm$0.11 \\
 & CRL-softmax & \color{red}\textbf{96.55$\pm$0.07} & 4.47$\pm$0.10 & 3.86$\pm$0.08 & 42.82$\pm$1.35 & 29.82$\pm$1.42 & \color{red}\textbf{0.88$\pm$0.12} & \color{red}\textbf{1.52$\pm$0.03} & \color{red}\textbf{5.44$\pm$0.10} \\
 & CRL-margin & 96.49$\pm$0.05 & 4.50$\pm$0.15 & 3.88$\pm$0.13 & 42.19$\pm$0.60 & 29.18$\pm$0.66 & 0.95$\pm$0.03 & 1.86$\pm$0.02 & 5.67$\pm$0.10 \\ \cline{2-10} 
 & MCdropout & 96.79$\pm$0.05 & 4.64$\pm$0.34 & 4.12$\pm$0.31 & 41.62$\pm$1.21 & 27.46$\pm$0.95 & \color{blue}\textbf{0.36$\pm$0.02} & \color{blue}\textbf{1.25$\pm$0.03} & \color{blue}\textbf{4.96$\pm$0.11} \\
 & Aleatoric+MC & \color{blue}\textbf{96.80$\pm$0.01} & 4.86$\pm$0.26 & 4.34$\pm$0.26 & 41.14$\pm$0.60 & 27.60$\pm$1.45 & 0.38$\pm$0.07 & 1.26$\pm$0.01 & 4.99$\pm$0.02 \\
 & AES(k=10) & 96.54$\pm$0.09 & 4.59$\pm$0.10 & 3.98$\pm$0.11 & 43.48$\pm$0.86 & 27.40$\pm$0.99 & 0.54$\pm$0.09 & 1.34$\pm$0.01 & 5.31$\pm$0.06 \\
 & AES(k=30) & 96.58$\pm$0.08 & \color{blue}\textbf{4.27$\pm$0.14} & \color{blue}\textbf{3.69$\pm$0.12} & \color{blue}\textbf{43.53$\pm$1.16} & \color{blue}\textbf{25.20$\pm$1.47} & 0.50$\pm$0.04 & 1.28$\pm$0.01 & 5.21$\pm$0.08 \\ \hline
\multirow{8}{*}{\begin{tabular}[c]{@{}l@{}}\textbf{SVHN}\\\textbf{ResNet110}\end{tabular}} & Baseline & 96.45$\pm$0.06 & 8.02$\pm$0.76 & 7.38$\pm$0.75 & 38.83$\pm$1.79 & 35.78$\pm$1.45 & 2.79$\pm$0.06 & 2.38$\pm$0.04 & 6.25$\pm$0.12 \\
 & CRL-entropy & 96.80$\pm$0.01 & 4.12$\pm$0.06 & 3.60$\pm$0.06 & 41.18$\pm$1.89 & 27.81$\pm$0.77 & 1.13$\pm$0.05 & 1.37$\pm$0.01 & 5.12$\pm$0.03 \\
 & CRL-softmax & 96.81$\pm$0.09 & 4.25$\pm$0.12 & 3.74$\pm$0.14 & \color{red}\textbf{43.46$\pm$1.78} & 27.71$\pm$0.56 & \color{red}\textbf{0.85$\pm$0.09} & \color{red}\textbf{1.31$\pm$0.02} & 4.97$\pm$0.12 \\
 & CRL-margin & \color{red}\textbf{96.83$\pm$0.09} & \color{red}\textbf{4.09$\pm$0.14} & \color{red}\textbf{3.58$\pm$0.15} & 42.32$\pm$2.42 & \color{red}\textbf{27.00$\pm$1.27} & 0.86$\pm$0.06 & 1.36$\pm$0.02 & \color{red}\textbf{4.93$\pm$0.08} \\ \cline{2-10} 
 & MCdropout & 97.00$\pm$0.00 & 4.99$\pm$0.35 & 4.53$\pm$0.34 & 39.10$\pm$0.94 & 28.69$\pm$2.22 & 0.65$\pm$0.07 & 1.29$\pm$0.01 & 4.73$\pm$0.13 \\
 & Aleatoric+MC & \color{blue}\textbf{97.01$\pm$0.04} & 5.54$\pm$0.24 & 5.09$\pm$0.23 & 38.71$\pm$1.08 & 31.60$\pm$0.50 & 0.54$\pm$0.05 & 1.25$\pm$0.01 & \color{blue}\textbf{4.69$\pm$0.05} \\
 & AES(k=10) & 96.77$\pm$0.05 & 4.41$\pm$0.17 & 3.89$\pm$0.16 & 43.56$\pm$2.51 & 27.39$\pm$1.34 & 0.43$\pm$0.11 & 1.26$\pm$0.01 & 4.97$\pm$0.05 \\
 & AES(k=30) & 96.81$\pm$0.05 & \color{blue}\textbf{4.23$\pm$0.13} & \color{blue}\textbf{3.72$\pm$0.14} & \color{blue}\textbf{43.64$\pm$1.48} & \color{blue}\textbf{26.09$\pm$1.54} & \color{blue}\textbf{0.33$\pm$0.03} & \color{blue}\textbf{1.21$\pm$0.02} & 4.89$\pm$0.05 \\ \hline
\multirow{8}{*}{\begin{tabular}[c]{@{}l@{}}\textbf{SVHN}\\\textbf{DenseNet}\end{tabular}} & Baseline & 96.40$\pm$0.08 & 7.70$\pm$0.41 & 7.00$\pm$0.39 & 39.43$\pm$0.78 & 34.23$\pm$1.21 & 2.51$\pm$0.07 & 2.10$\pm$0.05 & 6.13$\pm$0.15 \\
 & CRL-entropy & \color{red}\textbf{96.68$\pm$0.07} & \color{red}\textbf{4.27$\pm$0.34} & \color{red}\textbf{3.72$\pm$0.33} & 42.08$\pm$2.15 & 28.76$\pm$1.58 & 0.84$\pm$0.05 & 1.37$\pm$0.02 & 5.20$\pm$0.08 \\
 & CRL-softmax & 96.61$\pm$0.12 & 4.47$\pm$0.14 & 3.89$\pm$0.13 & \color{red}\textbf{43.35$\pm$0.81} & 28.35$\pm$1.62 & 0.85$\pm$0.06 & 1.38$\pm$0.04 & 5.26$\pm$0.18 \\
 & CRL-margin & 96.65$\pm$0.07 & 4.41$\pm$0.20 & 3.85$\pm$0.18 & 42.91$\pm$0.99 & \color{red}\textbf{26.58$\pm$1.04} & \color{red}\textbf{0.83$\pm$0.05} & \color{red}\textbf{1.35$\pm$0.00} & \color{red}\textbf{5.15$\pm$0.07} \\ \cline{2-10} 
 & MCdropout & 96.82$\pm$0.04 & 5.10$\pm$0.52 & 4.59$\pm$0.51 & 39.57$\pm$2.58 & 31.04$\pm$1.67 & 0.42$\pm$0.06 & 1.29$\pm$0.03 & 4.97$\pm$0.11 \\
 & Aleatoric+MC & \color{blue}\textbf{96.86$\pm$0.14} & 5.68$\pm$1.19 & 5.18$\pm$1.15 & 39.09$\pm$2.28 & 31.43$\pm$3.61 & 0.79$\pm$0.87 & 1.44$\pm$0.35 & 5.18$\pm$1.15 \\
 & AES(k=10) & 96.78$\pm$0.08 & 4.50$\pm$0.16 & 3.98$\pm$0.15 & \color{blue}\textbf{43.43$\pm$1.39} & 26.16$\pm$1.17 & 0.41$\pm$0.09 & 1.24$\pm$0.02 & 4.96$\pm$0.10 \\
 & AES(k=30) & 96.80$\pm$0.07 & \color{blue}\textbf{4.29$\pm$0.14} & \color{blue}\textbf{3.77$\pm$0.13} & 43.14$\pm$1.30 & \color{blue}\textbf{25.86$\pm$0.84} & \color{blue}\textbf{0.34$\pm$0.07} & \color{blue}\textbf{1.21$\pm$0.02} & \color{blue}\textbf{4.90$\pm$0.10} \\ \cline{1-10} 
\end{tabular}
    }
\end{table*}

\begin{table*}[h]
    \caption{Comparison of ensembles of five classifiers. $\lambda$ is set to $0.5$ for CRL models. For each experiment, the best result is shown in boldface. AURC and E-AURC values are multiplied by $10^3$, and NLL are multiplied by $10$ for clarity. All remaining values are percentage.}
    \label{stable-clf-ensemble-weight0.5}
    \vskip 0.2in
    \centering
    \resizebox{0.8\textwidth}{!}{
            \begin{tabular}{ll|ccccc|ccc}
\hline
\multirow{2}{*}{\textbf{\begin{tabular}[c]{@{}l@{}}\textbf{Dataset}\\ \textbf{Model}\end{tabular}}} & \multirow{2}{*}{\textbf{\textbf{Method}}} & \multirow{2}{*}{\textbf{\begin{tabular}[c]{@{}c@{}}\textbf{\small{ACC}}\\ \small{($\uparrow$)}\end{tabular}}} & \multirow{2}{*}{\textbf{\begin{tabular}[c]{@{}c@{}}\textbf{\small{AURC}}\\ \small{($\downarrow$)}\end{tabular}}} & \multirow{2}{*}{\textbf{\begin{tabular}[c]{@{}c@{}}\textbf{\small{E-AURC}}\\ \small{($\downarrow$)}\end{tabular}}} & \multirow{2}{*}{\textbf{\begin{tabular}[c]{@{}c@{}}\textbf{\small{AUPR-}}\\ \textbf{\small{Err}} \footnotesize{($\uparrow$)}\end{tabular}}} & \multirow{2}{*}{\textbf{\begin{tabular}[c]{@{}c@{}}\textbf{\small{FPR-95\%}}\\ \textbf{\small{TPR}} \small{($\downarrow$)}\end{tabular}}} & \multirow{2}{*}{\begin{tabular}[c]{@{}c@{}}\textbf{\small{ECE}}\\ \small{($\downarrow$)}\end{tabular}} & \multirow{2}{*}{\begin{tabular}[c]{@{}c@{}}\textbf{\small{NLL}}\\ \small{($\downarrow$)}\end{tabular}} & \multirow{2}{*}{\begin{tabular}[c]{@{}c@{}}\textbf{\small{Brier}}\\ \small{($\downarrow$)}\end{tabular}} \\
 &  &  &  &  &  &  &  &  &  \\ \hline
\multirow{4}{*}{\begin{tabular}[l]{@{}l@{}}\textbf{CIFAR-10}\\\textbf{VGG-16}\end{tabular}} & \small{Baseline} & 95.02 & 4.45 & 3.19 & \textbf{46.45} & \textbf{33.73} & 1.52 & 1.92 & 7.65 \\
 & \small{CRL-entropy} & 94.81 & 5.06 & 3.69 & 45.96 & 34.68 & 0.97 & 1.79 & 7.77 \\
 & \small{CRL-softmax} & \textbf{95.09} & \textbf{4.32} & \textbf{3.09} & 45.27 & 37.88 & 1.32 & 1.78 & \textbf{7.51} \\
 & \small{CRL-margin} & 94.85 & 5.05 & 3.70 & 42.01 & 40.77 & \textbf{0.93} & \textbf{1.71} & 7.67 \\ \hline
\multirow{4}{*}{\begin{tabular}[c]{@{}l@{}}\textbf{CIFAR-10}\\ \textbf{ResNet110}\end{tabular}} & \small{Baseline} & 95.42 & 4.01 & 2.95 & \textbf{44.14} & \textbf{29.03} & 1.12 & 1.63 & 6.86 \\
 & \small{CRL-entropy} & 95.15 & 4.12 & 2.93 & 43.38 & 34.02 & \textbf{0.42} & 1.50 & 7.22 \\
 & \small{CRL-softmax} & \textbf{95.55} & \textbf{3.72} & \textbf{2.72} & 44.01 & 29.88 & 0.84 & 1.50 & \textbf{6.60} \\
 & \small{CRL-margin} & 95.23 & 4.26 & 3.10 & 37.90 & 39.83 & 0.76 & \textbf{1.46} & 7.03 \\ \hline
\multirow{4}{*}{\begin{tabular}[c]{@{}l@{}}\textbf{CIFAR-10}\\ \textbf{DenseNet}\end{tabular}} & \small{Baseline} & \textbf{96.03} & \textbf{3.02} & \textbf{2.22} & 44.17 & 30.73 & 0.79 & 1.29 & \textbf{5.97} \\
 & \small{CRL-entropy} & 95.89 & 3.33 & 2.47 & 42.80 & 33.57 & 0.57 & 1.32 & 6.31 \\
 & \small{CRL-softmax} & 95.97 & 3.17 & 2.35 & 45.25 & 29.77 & 0.85 & \textbf{1.27} & 5.99 \\
 & \small{CRL-margin} & 95.50 & 3.45 & 2.43 & \textbf{47.12} & \textbf{28.88} & \textbf{0.45} & 1.32 & 6.48 \\ \hline
\multirow{4}{*}{\begin{tabular}[c]{@{}l@{}}\textbf{CIFAR-100}\\ \textbf{VGG-16}\end{tabular}} & \small{Baseline} & 78.34 & 54.53 & 29.16 & 64.99 & 58.44 & 4.07 & 9.53 & 31.05 \\
 & \small{CRL-entropy} & 78.43 & 55.19 & 30.05 & 64.50 & 60.36 & 3.85 & 9.14 & 30.86 \\
 & \small{CRL-softmax} & \textbf{78.53} & \textbf{52.53} & \textbf{27.63} & \textbf{66.53} & \textbf{57.89} & \textbf{3.80} & 9.11 & \textbf{30.47} \\
 & \small{CRL-margin} & 77.84 & 58.27 & 31.67 & 61.69 & 63.94 & 4.42 & \textbf{9.08} & 30.84 \\ \hline
\multirow{4}{*}{\begin{tabular}[c]{@{}l@{}}\textbf{CIFAR-100}\\ \textbf{ResNet110}\end{tabular}} & \small{Baseline} & 78.83 & 54.91 & 30.72 & 64.42 & 58.99 & 2.39 & 8.63 & 30.19 \\
 & \small{CRL-entropy} & 78.69 & 54.49 & 29.97 & 64.51 & 58.51 & \textbf{1.95} & 8.31 & 30.01 \\
 & \small{CRL-softmax} & \textbf{79.08} & \textbf{52.87} & \textbf{29.27} & \textbf{64.88} & \textbf{57.74} & 2.11 & \textbf{8.06} & \textbf{29.59} \\
 & \small{CRL-margin} & 79.01 & 57.20 & 33.44 & 56.87 & 68.41 & 2.04 & \textbf{8.06} & 29.90 \\ \hline
\multirow{4}{*}{\begin{tabular}[c]{@{}l@{}}\textbf{CIFAR-100}\\ \textbf{DenseNet}\end{tabular}} & \small{Baseline} & 80.34 & 47.43 & 26.70 & \textbf{63.83} & 56.10 & 1.87 & 7.43 & 27.74 \\
 & \small{CRL-entropy} & 80.47 & 46.10 & \textbf{25.65} & 63.73 & \textbf{55.65} & 1.81 & 7.20 & 27.47 \\
 & \small{CRL-softmax} & \textbf{80.85} & \textbf{45.63} & 25.99 & 61.46 & 57.33 & 1.79 & \textbf{7.13} & \textbf{27.34} \\
 & \small{CRL-margin} & 80.29 & 48.15 & 27.30 & 59.93 & 63.01 & \textbf{1.53} & 7.20 & 27.60 \\ \hline
\multirow{4}{*}{\begin{tabular}[c]{@{}l@{}}\textbf{SVHN}\\ \textbf{VGG-16}\end{tabular}} & \small{Baseline} & 96.91 & 4.48 & 4.00 & \textbf{40.66} & 28.64 & 1.09 & 1.60 & 4.93 \\
 & \small{CRL-entropy} & \textbf{97.01} & \textbf{3.96} & \textbf{3.51} & 39.80 & \textbf{27.02} & \textbf{0.78} & \textbf{1.30} & \textbf{4.75} \\
 & \small{CRL-softmax} & 96.95 & 4.07 & 3.60 & 40.52 & 29.25 & 1.02 & 1.53 & 4.92 \\
 & \small{CRL-margin} & 96.84 & 4.30 & 3.80 & 37.62 & 30.04 & 0.86 & 1.42 & 4.92 \\ \hline
\multirow{4}{*}{\begin{tabular}[c]{@{}l@{}}\textbf{SVHN}\\ \textbf{ResNet110}\end{tabular}} & \small{Baseline} & 97.13 & 4.33 & 3.91 & \textbf{42.52} & 26.30 & 0.92 & 1.38 & 4.47 \\
 & \small{CRL-entropy} & 97.24 & \textbf{3.56} & \textbf{3.17} & 41.58 & \textbf{25.80} & \textbf{0.59} & \textbf{1.13} & 4.30 \\
 & \small{CRL-softmax} & 97.29 & 3.80 & 3.43 & 40.75 & 26.80 & 0.88 & 1.23 & 4.26 \\
 & \small{CRL-margin} & \textbf{97.31} & 3.61 & 3.24 & 36.75 & 27.03 & 0.72 & 1.16 & \textbf{4.24} \\ \hline
\multirow{4}{*}{\begin{tabular}[c]{@{}l@{}}\textbf{SVHN}\\ \textbf{DenseNet}\end{tabular}} & \small{Baseline} & \textbf{97.24} & 4.93 & 4.55 & 36.49 & 30.54 & 0.83 & 1.34 & 4.51 \\
 & \small{CRL-entropy} & 97.15 & 3.85 & 3.44 & 40.59 & \textbf{27.16} & 0.72 & \textbf{1.17} & 4.47 \\
 & \small{CRL-softmax} & 97.18 & 4.10 & 3.70 & \textbf{43.31} & 29.05 & 0.87 & 1.25 & 4.46 \\
 & \small{CRL-margin} & 97.19 & \textbf{3.73} & \textbf{3.34} & 35.40 & 27.98 & \textbf{0.59} & 1.18 & \textbf{4.41} \\ \cline{1-10} 
\end{tabular}
    }
\end{table*}

\begin{table*}[h]
    \caption{Comparison of ensembles of five classifiers. $\lambda$ is set to $1$ for CRL models. For each experiment, the best result is shown in boldface. AURC and E-AURC values are multiplied by $10^3$, and NLL are multiplied by $10$ for clarity. All remaining values are percentage.}
    \label{stable-clf-ensemble-weight1}
    \vskip 0.2in
    \centering
    \resizebox{0.8\textwidth}{!}{%
        \begin{tabular}{ll|ccccc|ccc}
\hline
\multirow{2}{*}{\textbf{\begin{tabular}[c]{@{}l@{}}\textbf{Dataset}\\ \textbf{Model}\end{tabular}}} & \multirow{2}{*}{\textbf{\textbf{Method}}} & \multirow{2}{*}{\textbf{\begin{tabular}[c]{@{}c@{}}\textbf{\small{ACC}}\\ \small{($\uparrow$)}\end{tabular}}} & \multirow{2}{*}{\textbf{\begin{tabular}[c]{@{}c@{}}\textbf{\small{AURC}}\\ \small{($\downarrow$)}\end{tabular}}} & \multirow{2}{*}{\textbf{\begin{tabular}[c]{@{}c@{}}\textbf{\small{E-AURC}}\\ \small{($\downarrow$)}\end{tabular}}} & \multirow{2}{*}{\textbf{\begin{tabular}[c]{@{}c@{}}\textbf{\small{AUPR-}}\\ \textbf{\small{Err}} \footnotesize{($\uparrow$)}\end{tabular}}} & \multirow{2}{*}{\textbf{\begin{tabular}[c]{@{}c@{}}\textbf{\small{FPR-95\%}}\\ \textbf{\small{TPR}} \small{($\downarrow$)}\end{tabular}}} & \multirow{2}{*}{\begin{tabular}[c]{@{}c@{}}\textbf{\small{ECE}}\\ \small{($\downarrow$)}\end{tabular}} & \multirow{2}{*}{\begin{tabular}[c]{@{}c@{}}\textbf{\small{NLL}}\\ \small{($\downarrow$)}\end{tabular}} & \multirow{2}{*}{\begin{tabular}[c]{@{}c@{}}\textbf{\small{Brier}}\\ \small{($\downarrow$)}\end{tabular}} \\
 &  &  &  &  &  &  &  &  &  \\ \hline
\multirow{4}{*}{\begin{tabular}[c]{@{}l@{}}\textbf{CIFAR-10}\\ \textbf{VGG-16}\end{tabular}} & \small{Baseline} & \textbf{95.02} & \textbf{4.45} & \textbf{3.19} & 46.45 & \textbf{33.73} & 1.52 & 1.92 & \textbf{7.65} \\
 & \small{CRL-entropy} & 94.70 & 5.12 & 3.69 & 43.88 & 37.92 & \textbf{0.50} & 1.86 & 7.77 \\
 & \small{CRL-softmax} & 94.60 & 5.21 & 3.72 & \textbf{46.80} & 37.22 & 1.32 & \textbf{1.71} & 8.03 \\
 & \small{CRL-margin} & 94.77 & 5.67 & 4.28 & 36.91 & 47.22 & 0.99 & 1.90 & 8.15 \\ \hline
\multirow{4}{*}{\begin{tabular}[c]{@{}l@{}}\textbf{CIFAR-10}\\ \textbf{ResNet110}\end{tabular}} & \small{Baseline} & \textbf{95.42} & \textbf{4.01} & \textbf{2.95} & 44.14 & \textbf{29.03} & 1.12 & 1.63 & \textbf{6.86} \\
 & \small{CRL-entropy} & 95.16 & 4.42 & 3.23 & 39.56 & 35.95 & 1.68 & 1.63 & 7.43 \\
 & \small{CRL-softmax} & 94.70 & 4.58 & 3.15 & \textbf{45.23} & 34.15 & 0.72 & \textbf{1.53} & 7.71 \\
 & \small{CRL-margin} & 94.62 & 4.91 & 3.44 & 41.74 & 35.50 & \textbf{0.68} & 1.58 & 7.87 \\ \hline
\multirow{4}{*}{\begin{tabular}[c]{@{}l@{}}\textbf{CIFAR-10}\\ \textbf{DenseNet}\end{tabular}} & \small{Baseline} & \textbf{96.03} & \textbf{3.02} & \textbf{2.22} & \textbf{44.17} & \textbf{30.73} & 0.79 & \textbf{1.29} & \textbf{5.97} \\
 & \small{CRL-entropy} & 95.52 & 3.72 & 2.70 & 43.82 & 32.14 & 1.50 & 1.45 & 6.73 \\
 & \small{CRL-softmax} & 95.34 & 3.92 & 2.81 & 43.89 & 32.61 & \textbf{0.52} & 1.40 & 6.94 \\
 & \small{CRL-margin} & 95.18 & 4.26 & 3.08 & 40.61 & 37.75 & 0.61 & 1.45 & 7.28 \\ \hline
\multirow{4}{*}{\begin{tabular}[c]{@{}l@{}}\textbf{CIFAR-100}\\ \textbf{VGG-16}\end{tabular}} & \small{Baseline} & 78.34 & 54.53 & 29.16 & 64.99 & 58.44 & 4.07 & 9.53 & 31.05 \\
 & \small{CRL-entropy} & \textbf{78.66} & 55.05 & 28.46 & 65.20 & 59.04 & 2.17 & 8.59 & \textbf{29.96} \\
 & \small{CRL-softmax} & 78.09 & \textbf{53.74} & \textbf{27.76} & \textbf{67.01} & \textbf{56.86} & 2.76 & \textbf{8.48} & 30.29 \\
 & \small{CRL-margin} & 78.08 & 58.63 & 32.63 & 62.32 & 62.04 & \textbf{2.14} & 8.67 & 30.52 \\ \hline
\multirow{4}{*}{\begin{tabular}[c]{@{}l@{}}\textbf{CIFAR-100}\\ \textbf{ResNet110}\end{tabular}} & \small{Baseline} & 78.83 & 54.91 & 30.72 & 64.42 & 58.99 & 2.39 & 8.63 & 30.19 \\
 & \small{CRL-entropy} & 78.56 & 53.92 & 29.09 & 64.32 & 58.53 & 2.39 & 8.63 & 30.19 \\
 & \small{CRL-softmax} & 78.40 & \textbf{53.55} & \textbf{28.33} & \textbf{66.35} & \textbf{56.43} & 2.38 & 7.93 & 30.04 \\
 & \small{CRL-margin} & \textbf{78.84} & 55.85 & 31.69 & 58.53 & 66.82 & \textbf{1.78} & \textbf{7.61} & \textbf{29.69} \\ \hline
\multirow{4}{*}{\begin{tabular}[c]{@{}l@{}}\textbf{CIFAR-100}\\ \textbf{DenseNet}\end{tabular}} & \small{Baseline} & 80.34 & 47.43 & 26.70 & \textbf{63.83} & \textbf{56.10} & 1.87 & 7.43 & 27.74 \\
 & \small{CRL-entropy} & 80.18 & 47.37 & 26.29 & 62.65 & 56.91 & 2.18 & 7.21 & 27.73 \\
 & \small{CRL-softmax} & 80.38 & \textbf{46.63} & \textbf{25.98} & 62.59 & 58.81 & \textbf{1.45} & 6.95 & 27.43 \\
 & \small{CRL-margin} & \textbf{80.50} & 48.27 & 27.88 & 57.82 & 63.64 & 1.55 & \textbf{6.94} & \textbf{27.42} \\ \hline
\multirow{4}{*}{\begin{tabular}[c]{@{}l@{}}\textbf{SVHN}\\ \textbf{VGG-16}\end{tabular}} & \small{Baseline} & 96.91 & 4.48 & 4.00 & 40.66 & 28.64 & 1.09 & 1.60 & 4.93 \\
 & \small{CRL-entropy} & \textbf{96.98} & 4.16 & 3.70 & \textbf{41.49} & \textbf{26.62} & \textbf{0.45} & \textbf{1.30} & \textbf{4.75} \\
 & \small{CRL-softmax} & \textbf{96.98} & \textbf{4.02} & \textbf{3.56} & 41.21 & 28.95 & 0.81 & \textbf{1.30} & 4.79 \\
 & \small{CRL-margin} & 96.97 & 4.05 & 3.59 & 38.50 & 29.18 & 0.47 & 1.46 & 4.87 \\ \hline
\multirow{4}{*}{\begin{tabular}[c]{@{}l@{}}\textbf{SVHN}\\ \textbf{ResNet110}\end{tabular}} & \small{Baseline} & 97.13 & 4.33 & 3.91 & \textbf{42.52} & 26.30 & 0.92 & 1.38 & 4.47 \\
 & \small{CRL-entropy} & \textbf{97.31} & \textbf{3.51} & \textbf{3.15} & 37.65 & 28.08 & 0.60 & 1.13 & 4.30 \\
 & \small{CRL-softmax} & 97.26 & 3.82 & 3.44 & 40.00 & 26.58 & 0.56 & \textbf{1.12} & 4.33 \\
 & \small{CRL-margin} & 97.26 & 3.66 & 3.28 & 37.61 & \textbf{25.17} & \textbf{0.50} & 1.14 & \textbf{4.27} \\ \hline
\multirow{4}{*}{\begin{tabular}[c]{@{}l@{}}\textbf{SVHN}\\ \textbf{DenseNet}\end{tabular}} & \small{Baseline} & \textbf{97.24} & 4.93 & 4.55 & 36.49 & 30.54 & 0.83 & 1.34 & 4.51 \\
 & \small{CRL-entropy} & 97.18 & \textbf{3.70} & \textbf{3.30} & 39.74 & 26.43 & 0.74 & 1.16 & \textbf{4.44} \\
 & \small{CRL-softmax} & 97.13 & 3.85 & 3.44 & 39.91 & \textbf{25.77} & \textbf{0.53} & \textbf{1.14} & 4.46 \\
 & \small{CRL-margin} & 97.19 & 3.76 & 3.37 & \textbf{40.02} & 28.49 & 0.81 & 1.17 & 4.53 \\ \cline{1-10} 
\end{tabular}
    }
\end{table*}

\begingroup
\renewcommand{\arraystretch}{1.2} 
\begin{table*}[h]
\centering
\caption{Performances of CRL models on out-of-distribution detection task. The means and standard deviations are computed from five models trained to evaluate the ordinal ranking performance. For each comparison, better result is shown in boldface. All values are percentage.}
\label{stable-ood-perf}
\vskip 0.1in
\resizebox{\textwidth}{!}{
\begin{tabular}{llccccc} \toprule
\multicolumn{1}{l}{\begin{tabular}[c]{@{}l@{}}\textbf{In-dist}\\\textbf{Model} \end{tabular}} & \multicolumn{1}{l}{\textbf{Out-of-dist} } & \textbf{FPR-95\%TPR($\downarrow$)}  & \textbf{Detection Err.($\downarrow$)}  & \textbf{AUROC($\uparrow$)}  & \textbf{AUPR-In($\uparrow$)}  & \textbf{AUPR-Out($\uparrow$)}  \\ \hline
\multicolumn{1}{c}{\textbf{} } & \multicolumn{1}{c}{} & \multicolumn{5}{c}{\begin{tabular}[c]{@{}c@{}}\textbf{Baseline / CRL}\\\textbf{ Baseline+ODIN / CRL+ODIN }\\\textbf{ Baseline+Mahalanobis / CRL+Mahalanobis } \end{tabular}} \\ \cline{3-7}
\multirow{7.3}{*}{\begin{tabular}[c]{@{}l@{}}\textbf{SVHN}\\\textbf{ResNet110} \end{tabular}} & TinyImageNet & \begin{tabular}[c]{@{}c@{}} 29.65$\pm$2.40 / \textbf{5.89$\pm$0.70\;\;} \\ 27.50$\pm$3.09 / \textbf{2.17$\pm$0.26\;\;} \\ \textbf{0.24$\pm$0.08} / 0.30$\pm$0.12 \end{tabular} & \begin{tabular}[c]{@{}c@{}} 12.11$\pm$0.96 / \textbf{5.05$\pm$0.23\;\;} \\ 13.14$\pm$1.29 / \textbf{3.41$\pm$0.23\;\;} \\ \textbf{1.15$\pm$0.16} / 1.39$\pm$0.27 \end{tabular} & \begin{tabular}[c]{@{}c@{}} 93.00$\pm$1.06 / \textbf{98.83$\pm$0.13}\\ 92.32$\pm$1.38 / \textbf{99.39$\pm$0.08} \\ \textbf{99.88$\pm$0.03} / 99.82$\pm$0.06 \end{tabular} & \begin{tabular}[c]{@{}c@{}} 96.31$\pm$0.91 / \textbf{99.56$\pm$0.04}\\95.63$\pm$1.17 / \textbf{99.76$\pm$0.03} \\ \textbf{99.96$\pm$0.01} / 99.93$\pm$0.02 \end{tabular} & \begin{tabular}[c]{@{}c@{}} 84.95$\pm$1.41 / \textbf{96.72$\pm$0.50}\\ 85.61$\pm$1.85 / \textbf{98.41$\pm$0.30} \\ \textbf{99.39$\pm$0.19} / 98.99$\pm$0.28 \end{tabular} \\ \cline{3-7}
 & LSUN & \begin{tabular}[c]{@{}c@{}} 32.37$\pm$2.78 / \textbf{7.48$\pm$0.91\;\;}\\ 29.57$\pm$3.98 / \textbf{2.92$\pm$0.51\;\;} \\ 0.08$\pm$0.05 / \textbf{0.06$\pm$0.07} \end{tabular} & \begin{tabular}[c]{@{}c@{}} 13.01$\pm$1.17 / \textbf{5.50$\pm$0.20}\;\;\\ 14.06$\pm$1.23 / \textbf{3.88$\pm$0.30}\;\; \\ 0.88$\pm$0.14 / \textbf{0.85$\pm$0.32} \end{tabular} & \begin{tabular}[c]{@{}c@{}} 92.19$\pm$1.39 / \textbf{98.62$\pm$0.17}\\ 91.56$\pm$1.78 / \textbf{99.28$\pm$0.11} \\ \textbf{99.91$\pm$0.03} / 99.89$\pm$0.06 \end{tabular} & \begin{tabular}[c]{@{}c@{}} 95.82$\pm$1.30 / \textbf{99.49$\pm$0.06}\\ 95.19$\pm$1.65 / \textbf{99.72$\pm$0.03} \\ \textbf{99.97$\pm$0.01} / 99.96$\pm$0.02 \end{tabular} & \begin{tabular}[c]{@{}c@{}} 83.48$\pm$1.74 / \textbf{96.14$\pm$0.62} \\ 84.42$\pm$2.38 / \textbf{98.06$\pm$0.42} \\ \textbf{99.45$\pm$0.25} / 99.03$\pm$0.38 \end{tabular} \\ \cline{3-7}
 & iSUN & \begin{tabular}[c]{@{}c@{}} 31.43$\pm$2.63 / \textbf{6.40$\pm$0.78\;\;}\\ 29.73$\pm$3.71 / \textbf{2.55$\pm$0.41\;\;} \\ 0.04$\pm$0.05 / \textbf{0.03$\pm$0.03} \end{tabular} & \begin{tabular}[c]{@{}c@{}} 12.67$\pm$1.15 / \textbf{5.17$\pm$0.23}\;\; \\ 13.87$\pm$1.43 / \textbf{3.67$\pm$0.26}\;\; \\ \textbf{0.73$\pm$0.22} / 0.81$\pm$0.24 \end{tabular} & \begin{tabular}[c]{@{}c@{}} 92.46$\pm$1.39 / \textbf{98.75$\pm$0.14}\\ 91.59$\pm$1.91 / \textbf{99.30$\pm$0.07} \\ \textbf{99.89$\pm$0.03} / 99.88$\pm$0.04 \end{tabular} & \begin{tabular}[c]{@{}c@{}} 96.36$\pm$1.21 / \textbf{99.58$\pm$0.04}\\ 95.67$\pm$1.58 / \textbf{99.76$\pm$0.02} \\ 99.97$\pm$0.04 / \textbf{99.97$\pm$0.01} \end{tabular} & \begin{tabular}[c]{@{}c@{}} 82.54$\pm$1.66 / \textbf{96.13$\pm$0.59}\\ 83.04$\pm$2.42 / \textbf{97.92$\pm$0.29} \\ \textbf{99.31$\pm$0.26} / 98.79$\pm$0.63 \end{tabular} \\ \cline{1-7}
\multirow{7.3}{*}{\begin{tabular}[c]{@{}l@{}}\textbf{SVHN}\\\textbf{DenseNet} \end{tabular}} & TinyImageNet & \begin{tabular}[c]{@{}c@{}} 26.32$\pm$5.55 / \textbf{7.99$\pm$2.49\;\;}\\ 19.93$\pm$4.43 / \textbf{3.39$\pm$1.34\;\;}\\ 1.44$\pm$1.62 / \textbf{1.03$\pm$1.41} \end{tabular} & \begin{tabular}[c]{@{}c@{}} 11.49$\pm$1.61 / \textbf{5.75$\pm$0.71}\;\;\\ 11.46$\pm$1.80 / \textbf{4.04$\pm$0.80}\;\;\\ 2.42$\pm$1.15 / \textbf{1.86$\pm$0.92} \end{tabular} & \begin{tabular}[c]{@{}c@{}} 93.75$\pm$1.43 / \textbf{98.53$\pm$0.41}\\ 94.06$\pm$1.47 / \textbf{99.17$\pm$0.28} \\ 99.37$\pm$0.91 / \textbf{99.48$\pm$0.79} \end{tabular} & \begin{tabular}[c]{@{}c@{}} 96.62$\pm$0.94 / \textbf{99.43$\pm$0.16}\\ 96.56$\pm$0.94 / \textbf{99.65$\pm$0.12} \\ 99.52$\pm$1.08 / \textbf{99.62$\pm$0.99} \end{tabular} & \begin{tabular}[c]{@{}c@{}} 87.30$\pm$2.74 / \textbf{96.15$\pm$1.16}\\ 90.03$\pm$2.39 / \textbf{98.00$\pm$0.67} \\ 98.45$\pm$1.04 / \textbf{98.48$\pm$0.80} \end{tabular} \\ \cline{3-7}
 & LSUN & \begin{tabular}[c]{@{}c@{}} 28.95$\pm$5.80 / \textbf{11.05$\pm$3.09}\\ 22.22$\pm$4.86 / \textbf{4.63$\pm$1.84\;\;} \\ \textbf{0.41$\pm$0.84} / 0.44$\pm$0.46 \end{tabular} & \begin{tabular}[c]{@{}c@{}} 12.39$\pm$1.84 / \textbf{6.58$\pm$0.74}\;\;\\ 12.35$\pm$1.98 / \textbf{4.68$\pm$0.94}\;\; \\\textbf{1.23$\pm$0.63} / 1.23$\pm$0.66 \end{tabular} & \begin{tabular}[c]{@{}c@{}} 92.95$\pm$1.76 / \textbf{98.12$\pm$0.49}\\ 93.32$\pm$1.80 / \textbf{98.93$\pm$0.36} \\ 99.73$\pm$0.60 / \textbf{99.75$\pm$0.17} \end{tabular} & \begin{tabular}[c]{@{}c@{}} 96.11$\pm$1.23 / \textbf{99.29$\pm$0.19}\\ 96.15$\pm$1.22 / \textbf{99.56$\pm$0.15} \\ \textbf{99.86$\pm$0.88} / 99.79$\pm$0.13 \end{tabular} & \begin{tabular}[c]{@{}c@{}} 85.93$\pm$3.12 / \textbf{95.06$\pm$1.40}\\ 88.83$\pm$2.73 / \textbf{97.45$\pm$0.89} \\ \textbf{98.97$\pm$0.60} / 98.70$\pm$0.62 \end{tabular} \\ \cline{3-7}
 & iSUN & \begin{tabular}[c]{@{}c@{}} 28.22$\pm$5.59 / \textbf{9.94$\pm$2.78\;\;}\\ 22.37$\pm$4.59 / \textbf{4.39$\pm$1.90\;\;} \\ \textbf{0.06$\pm$0.46} / 0.08$\pm$0.16 \end{tabular} & \begin{tabular}[c]{@{}c@{}} 12.08$\pm$1.62 / \textbf{6.43$\pm$0.73}\;\;\\ 12.20$\pm$1.82 / \textbf{4.51$\pm$1.03}\;\; \\ \textbf{0.81$\pm$0.62} / 1.13$\pm$0.26 \end{tabular} & \begin{tabular}[c]{@{}c@{}} 93.11$\pm$1.57 / \textbf{98.25$\pm$0.47}\\ 93.37$\pm$1.62 / \textbf{98.96$\pm$0.36} \\ \textbf{99.87$\pm$0.21} / 99.78$\pm$0.05 \end{tabular} & \begin{tabular}[c]{@{}c@{}} 96.57$\pm$0.96 / \textbf{99.39$\pm$0.16}\\ 96.51$\pm$0.97 / \textbf{99.61$\pm$0.14} \\ 99.94$\pm$0.17 / \textbf{99.94$\pm$0.03} \end{tabular} & \begin{tabular}[c]{@{}c@{}} 85.04$\pm$3.20 / \textbf{95.04$\pm$1.50}\\ 87.95$\pm$2.78 / \textbf{97.38$\pm$0.99} \\ \textbf{98.92$\pm$0.33} / 98.56$\pm$0.30 \end{tabular} \\ \cline{1-7}
\multirow{7}{*}{\begin{tabular}[c]{@{}l@{}}\textbf{CIFAR-10}\\\textbf{ResNet110} \end{tabular}} & TinyImageNet & \begin{tabular}[c]{@{}c@{}} 66.09$\pm$2.86 / \textbf{53.17$\pm$5.60}\\ 49.33$\pm$4.19 / \textbf{43.08$\pm$5.15} \\ \textbf{8.46$\pm$2.12} / 9.44$\pm$2.25 \end{tabular} & \begin{tabular}[c]{@{}c@{}} 22.59$\pm$1.81 / \textbf{22.06$\pm$2.35}\\ 22.08$\pm$2.28 / \textbf{17.69$\pm$2.02} \\ \textbf{6.39$\pm$0.87} / 7.02$\pm$0.91 \end{tabular} & \begin{tabular}[c]{@{}c@{}} 82.59$\pm$2.91 / \textbf{86.25$\pm$2.76}\\ 84.31$\pm$3.22 / \textbf{90.40$\pm$1.91} \\ \textbf{98.34$\pm$0.41} / 97.92$\pm$0.41 \end{tabular} & \begin{tabular}[c]{@{}c@{}} 79.63$\pm$5.39 / \textbf{86.56$\pm$3.27}\\ 80.73$\pm$5.07 / \textbf{90.77$\pm$2.13} \\ \textbf{98.40$\pm$0.37} / 97.85$\pm$0.35 \end{tabular} & \begin{tabular}[c]{@{}c@{}} 82.07$\pm$2.00 / \textbf{85.61$\pm$2.50}\\ 86.01$\pm$2.30 / \textbf{90.03$\pm$1.77} \\ \textbf{98.22$\pm$0.49} / 98.02$\pm$0.44 \end{tabular} \\ \cline{3-7}
 & LSUN & \begin{tabular}[c]{@{}c@{}} 57.65$\pm$2.89 / \textbf{44.53$\pm$6.57}\\ 34.72$\pm$5.75 / \textbf{32.10$\pm$5.29} \\ 6.33$\pm$2.54 / \textbf{5.52$\pm$1.39} \end{tabular} & \begin{tabular}[c]{@{}c@{}} \textbf{17.78$\pm$1.21} / 17.89$\pm$1.87\\ 16.29$\pm$1.76 / \textbf{13.50$\pm$1.64} \\ 5.51$\pm$1.25 / \textbf{5.16$\pm$0.76} \end{tabular} & \begin{tabular}[c]{@{}c@{}} 88.25$\pm$1.54 / \textbf{90.46$\pm$1.93}\\ 90.63$\pm$1.97 / \textbf{93.90$\pm$1.23} \\ 98.66$\pm$0.51 / \textbf{98.71$\pm$0.29} \end{tabular} & \begin{tabular}[c]{@{}c@{}} 87.73$\pm$2.59 / \textbf{91.37$\pm$1.95}\\ 88.98$\pm$2.79 / \textbf{94.48$\pm$1.21} \\ \textbf{98.79$\pm$0.48} / 98.76$\pm$0.25 \end{tabular} & \begin{tabular}[c]{@{}c@{}} 87.06$\pm$1.29 / \textbf{89.60$\pm$1.97}\\ 91.47$\pm$1.66 / \textbf{93.30$\pm$1.25} \\ 98.49$\pm$0.61 / \textbf{98.62$\pm$0.31} \end{tabular} \\ \cline{3-7}
 & iSUN & \begin{tabular}[c]{@{}c@{}} 61.78$\pm$2.67 / \textbf{50.34$\pm$5.87}\\ 41.95$\pm$4.35 / \textbf{38.08$\pm$5.21} \\ \textbf{8.19$\pm$2.47} / 9.48$\pm$1.66 \end{tabular} & \begin{tabular}[c]{@{}c@{}} 20.07$\pm$1.47 / \textbf{20.01$\pm$2.12}\\ 19.02$\pm$1.70 / \textbf{14.89$\pm$1.58} \\ \textbf{6.53$\pm$1.08} / 7.07$\pm$0.71 \end{tabular} & \begin{tabular}[c]{@{}c@{}} 85.84$\pm$2.15 / \textbf{88.39$\pm$2.27}\\ 87.95$\pm$2.39 / \textbf{92.66$\pm$1.31} \\ \textbf{98.10$\pm$0.57} / 97.97$\pm$0.37 \end{tabular} & \begin{tabular}[c]{@{}c@{}} 85.78$\pm$3.71 / \textbf{90.25$\pm$2.28}\\ 86.90$\pm$3.59 / \textbf{93.96$\pm$1.17} \\ \textbf{98.29$\pm$0.50} / 98.10$\pm$0.33 \end{tabular} & \begin{tabular}[c]{@{}c@{}} 83.44$\pm$1.63 / \textbf{86.29$\pm$2.37}\\ 88.08$\pm$1.90 / \textbf{91.01$\pm$1.48} \\ \textbf{97.89$\pm$0.72} / 97.81$\pm$0.42 \end{tabular} \\ \cline{1-7}
\multirow{7}{*}{\begin{tabular}[c]{@{}l@{}}\textbf{CIFAR-10}\\\textbf{DenseNet}  \end{tabular}} & TinyImageNet & \begin{tabular}[c]{@{}c@{}} 45.81$\pm$3.95 / \textbf{29.87$\pm$4.09}\\ 10.73$\pm$6.24 / \textbf{10.41$\pm$3.09} \\ 6.99$\pm$1.13 / \textbf{6.28$\pm$3.18} \end{tabular} & \begin{tabular}[c]{@{}c@{}} 13.15$\pm$1.41 / \textbf{12.99$\pm$1.03}\\ 7.09$\pm$2.02 / \textbf{6.89$\pm$1.10} \\ 5.92$\pm$0.58 / \textbf{5.61$\pm$1.54} \end{tabular} & \begin{tabular}[c]{@{}c@{}} 93.25$\pm$1.04 / \textbf{94.50$\pm$0.84}\\ 97.86$\pm$1.09 / \textbf{97.97$\pm$0.56} \\ 98.37$\pm$0.50 / \textbf{98.52$\pm$1.17} \end{tabular} & \begin{tabular}[c]{@{}c@{}} 94.53$\pm$0.94 / \textbf{95.17$\pm$0.71}\\ 97.90$\pm$1.04 / \textbf{98.16$\pm$0.48} \\ \textbf{98.22$\pm$1.29} / 98.09$\pm$2.07 \end{tabular} & \begin{tabular}[c]{@{}c@{}} 91.82$\pm$1.24 / \textbf{93.87$\pm$1.03}\\ \textbf{97.84$\pm$1.13} / 97.78$\pm$0.65 \\ 98.49$\pm$0.38 / \textbf{98.57$\pm$0.82} \end{tabular} \\ \cline{3-7}
 & LSUN & \begin{tabular}[c]{@{}c@{}} 36.31$\pm$3.64 / \textbf{21.22$\pm$2.73}\\ \textbf{4.32$\pm$2.55} / 5.29$\pm$1.53 \\ 5.27$\pm$1.15 / \textbf{3.86$\pm$2.15} \end{tabular} & \begin{tabular}[c]{@{}c@{}} 10.60$\pm$0.88 / \textbf{10.59$\pm$0.55}\\ \textbf{4.46$\pm$1.15} / 5.03$\pm$0.71 \\ 5.08$\pm$0.57 / \textbf{4.26$\pm$1.11} \end{tabular} & \begin{tabular}[c]{@{}c@{}} 95.18$\pm$0.64 / \textbf{96.34$\pm$0.44}\\ \textbf{99.04$\pm$0.46} / 98.81$\pm$0.29 \\ 98.73$\pm$0.50 / \textbf{98.89$\pm$0.66} \end{tabular} & \begin{tabular}[c]{@{}c@{}} 96.16$\pm$0.50 / \textbf{96.80$\pm$0.35}\\ \textbf{99.10$\pm$0.41} / 98.92$\pm$0.24 \\ \textbf{98.68$\pm$1.56} / 98.67$\pm$1.07 \end{tabular} & \begin{tabular}[c]{@{}c@{}} 94.14$\pm$0.86 / \textbf{95.94$\pm$0.57}\\ \textbf{98.99$\pm$0.50} / 98.70$\pm$0.35 \\ 98.71$\pm$0.36 / \textbf{98.91$\pm$0.50} \end{tabular} \\ \cline{3-7}
 & iSUN & \begin{tabular}[c]{@{}c@{}} 39.84$\pm$4.40 / \textbf{25.59$\pm$3.52}\\ \textbf{6.61$\pm$4.00} / 7.54$\pm$2.08 \\ 6.49$\pm$1.07 / \textbf{6.20$\pm$1.45} \end{tabular} & \begin{tabular}[c]{@{}c@{}} \textbf{11.49$\pm$1.26} / 11.75$\pm$0.84\\ \textbf{5.49$\pm$1.41} / 6.03$\pm$0.92 \\ 5.67$\pm$0.52 / \textbf{5.54$\pm$0.69} \end{tabular} & \begin{tabular}[c]{@{}c@{}} 94.51$\pm$0.91 / \textbf{95.52$\pm$0.67}\\ \textbf{98.67$\pm$0.70} / 98.45$\pm$0.41 \\ \textbf{98.52$\pm$0.21} / 98.49$\pm$0.51 \end{tabular} & \begin{tabular}[c]{@{}c@{}} 95.99$\pm$0.70 / \textbf{96.44$\pm$0.50}\\ \textbf{98.87$\pm$0.55} / 98.73$\pm$0.33 \\ 98.34$\pm$0.30 / \textbf{98.38$\pm$0.93} \end{tabular} & \begin{tabular}[c]{@{}c@{}} 92.61$\pm$1.26 / \textbf{94.52$\pm$0.88}\\ \textbf{98.46$\pm$0.87} / 98.13$\pm$0.54 \\ 98.44$\pm$0.19 / \textbf{98.50$\pm$0.41} \end{tabular} \\ \bottomrule
\end{tabular}
}
\end{table*}
\endgroup

\begingroup
\renewcommand{\arraystretch}{1.2}
\begin{table*}[t]
\caption{Comparison of five sampling strategies for ResNet18 on CIFAR datasets. The means and standard deviations of accuracy values over five runs are reported. For each experiment, the best result is shown in boldface. The percentage in parentheses next to the stage number indicates the proportion of the labeled dataset to the entire training dataset (i.e., 50,000).}
\label{stable-al-perf}
\vskip 0.1in
\resizebox{\textwidth}{!}{
\begin{tabular}{ll|cccccccccc}
\hline
\multirow{2}{*}{\textbf{\textbf{Dataset}}} & \multirow{2}{*}{\textbf{\textbf{Sampling}}} & \multicolumn{10}{c}{\textbf{Stage}} \\
 &  & \textbf{1st} (4\%) & \textbf{2nd} (8\%) & \textbf{3rd} (12\%) & \textbf{4th} (16\%) & \textbf{5th} (20\%) & \textbf{6th} (24\%) & \textbf{7th} (28\%) & \textbf{8th} (32\%) & \textbf{9th} (36\%) & \textbf{10th} (40\%) \\ \hline
\multirow{5}{*}{\textbf{CIFAR-10}} & random & 64.86$\pm$1.43 & 77.35$\pm$0.69 & 82.31$\pm$1.42 & 85.43$\pm$0.96 & 86.84$\pm$0.71 & 88.36$\pm$0.82 & 89.45$\pm$0.70 & 90.47$\pm$0.54 & 91.10$\pm$0.57 & 91.50$\pm$0.58 \\
 & entropy & 64.98$\pm$1.12 & 80.40$\pm$0.56 & 85.69$\pm$0.59 & 88.30$\pm$0.34 & 90.14$\pm$0.28 & 91.62$\pm$0.19 & 92.69$\pm$0.13 & 93.35$\pm$0.15 & 93.82$\pm$0.26 & 94.32$\pm$0.18 \\
 & coreset & 65.36$\pm$0.95 & 79.55$\pm$0.39 & 84.79$\pm$0.23 & 87.58$\pm$0.24 & 89.42$\pm$0.19 & 91.00$\pm$0.24 & 91.94$\pm$0.34 & 92.70$\pm$0.21 & 93.48$\pm$0.16 & 93.81$\pm$0.20 \\
 & MC-entropy & 59.22$\pm$1.89 & 75.54$\pm$1.46 & 84.62$\pm$0.59 & 87.93$\pm$0.52 & 90.10$\pm$0.41 & 91.55$\pm$0.22 & 92.72$\pm$0.17 & 93.32$\pm$0.19 & \textbf{93.96$\pm$0.17} & \textbf{94.33$\pm$0.17} \\
 & CRL-softmax & \textbf{65.91$\pm$1.44} & \textbf{80.60$\pm$0.58} & \textbf{85.84$\pm$0.36} & \textbf{89.00$\pm$0.06} & \textbf{90.82$\pm$0.16} & \textbf{91.80$\pm$0.29} & \textbf{92.78$\pm$0.10} & \textbf{93.38$\pm$0.24} & 93.82$\pm$0.14 & 94.09$\pm$0.09 \\ \hline
\multirow{5}{*}{\textbf{CIFAR-100}} & random & 20.57$\pm$0.49 & 31.72$\pm$1.36 & 42.61$\pm$1.45 & 48.84$\pm$0.35 & 54.30$\pm$0.41 & 58.34$\pm$0.50 & 61.11$\pm$0.16 & 63.39$\pm$0.28 & 65.19$\pm$0.28 & 66.73$\pm$0.28 \\
 & entropy & 19.86$\pm$0.42 & 32.41$\pm$0.17 & 42.63$\pm$1.30 & 50.45$\pm$0.80 & 55.36$\pm$0.47 & 60.10$\pm$0.45 & 63.22$\pm$0.50 & 65.54$\pm$0.46 & 68.01$\pm$0.42 & 69.35$\pm$0.23 \\
 & coreset & 20.27$\pm$0.64 & \textbf{33.47$\pm$0.74} & \textbf{44.36$\pm$1.20} & 50.31$\pm$0.46 & 56.00$\pm$0.65 & 59.34$\pm$0.46 & 62.53$\pm$0.43 & 64.82$\pm$0.27 & 66.66$\pm$0.37 & 68.20$\pm$0.28 \\
 & MC-entropy & 19.45$\pm$0.70 & 31.20$\pm$2.18 & 41.57$\pm$1.18 & 50.03$\pm$0.69 & 56.18$\pm$0.60 & 60.28$\pm$0.34 & 63.55$\pm$0.20 & 66.31$\pm$0.22 & 68.25$\pm$0.35 & 69.97$\pm$0.40 \\
 & CRL-softmax & \textbf{20.72$\pm$0.34} & 32.17$\pm$1.03 & 43.87$\pm$1.82 & \textbf{51.22$\pm$0.99} & \textbf{57.69$\pm$0.79} & \textbf{61.65$\pm$0.31} & \textbf{64.27$\pm$0.36} & \textbf{66.71$\pm$0.57} & \textbf{68.78$\pm$0.27} & \textbf{70.40$\pm$0.32} \\ \cline{1-12} 
\end{tabular}
}
\end{table*}
\endgroup

\end{document}